\documentclass[final]{cvpr}
\usepackage[accsupp]{axessibility}
\usepackage{times}
\usepackage{epsfig}
\usepackage{graphicx}
\usepackage{amsmath}
\usepackage{amssymb}
\usepackage{amssymb}
\usepackage{bbm}
\DeclareMathAlphabet\mathbfcal{OMS}{cmsy}{b}{n}

\usepackage{float}

\usepackage{lipsum}
\usepackage{stfloats}
\usepackage{multicol}
\usepackage{multirow}
\usepackage{bm}
\usepackage{etoolbox}

\usepackage{array}
\usepackage{tabulary}
\usepackage[table]{xcolor} 
\usepackage{paralist}
\usepackage{booktabs}
\usepackage{adjustbox}

\usepackage{caption}
\usepackage{subcaption}

\usepackage[pagebackref=true,breaklinks=true,colorlinks,bookmarks=true,bookmarksopen=true]{hyperref}
\hypersetup{colorlinks, citecolor=blue}
\hypersetup{colorlinks, linkcolor=red}

\definecolor{lightred}{HTML}{F19C99}
\newcommand{\green}[1]{\textcolor[RGB]{96,177,87}{#1}}
\newcommand{\fn}[1]{{#1}}
\newcommand{\gbf}[1]{\green{\bf{\fn{(#1)}}}}

\definecolor{graytablerow}{gray}{0.6}
\newcommand{\grow}[1]{\textcolor{graytablerow}{#1}}

\def\cvprPaperID{2748} 
\def\confYear{CVPR 2022}
\begin{document}

\title{Bending Reality: Distortion-aware Transformers for Adapting to Panoramic Semantic Segmentation}
\author{Jiaming Zhang$^1$
~Kailun Yang$^{1}$\thanks{Corresponding author (e-mail: {\tt kailun.yang@kit.edu}).}
~Chaoxiang Ma$^2$
~Simon Reiß$^{1,3}$
~Kunyu Peng$^1$
~Rainer Stiefelhagen$^1$\\
\normalsize
$^1$CV:HCI Lab, Karlsruhe Institute of Technology
\normalsize
~$^2$ByteDance Inc.
\normalsize
~$^3$Carl Zeiss AG
}

\maketitle

\thispagestyle{empty}

\begin{abstract}
Panoramic images with their $360^\circ$ directional view encompass exhaustive information about the surrounding space, providing a rich foundation for scene understanding. To unfold this potential in the form of robust panoramic segmentation models, large quantities of expensive, pixel-wise annotations are crucial for success. Such annotations are available, but predominantly for narrow-angle, pinhole-camera images which, off the shelf, serve as sub-optimal resources for training panoramic models. Distortions and the distinct image-feature distribution in $360^\circ$ panoramas impede the transfer from the annotation-rich pinhole domain and therefore come with a big dent in performance. To get around this domain difference and bring together semantic annotations from pinhole- and $360^\circ$ surround-visuals, we propose to learn object deformations and panoramic image distortions in the Deformable Patch Embedding~(DPE) and Deformable MLP~(DMLP) components which blend into our \emph{Transformer for PAnoramic Semantic Segmentation~(Trans4PASS)} model. Finally, we tie together shared semantics in pinhole- and panoramic feature embeddings by generating multi-scale prototype features and aligning them in our Mutual Prototypical Adaptation (MPA) for unsupervised domain adaptation. On the indoor Stanford2D3D dataset, our Trans4PASS with MPA maintains comparable performance to fully-supervised state-of-the-arts, cutting the need for over $1,400$ labeled panoramas. On the outdoor DensePASS dataset, we break state-of-the-art by $14.39\%$ mIoU and set the new bar at $56.38\%$.\footnote[1]{Code will be made publicly available at \url{https://github.com/jamycheung/Trans4PASS}.} 

\end{abstract}

\begin{figure}[!t]
	\centering
    \includegraphics[width=1.0\columnwidth]{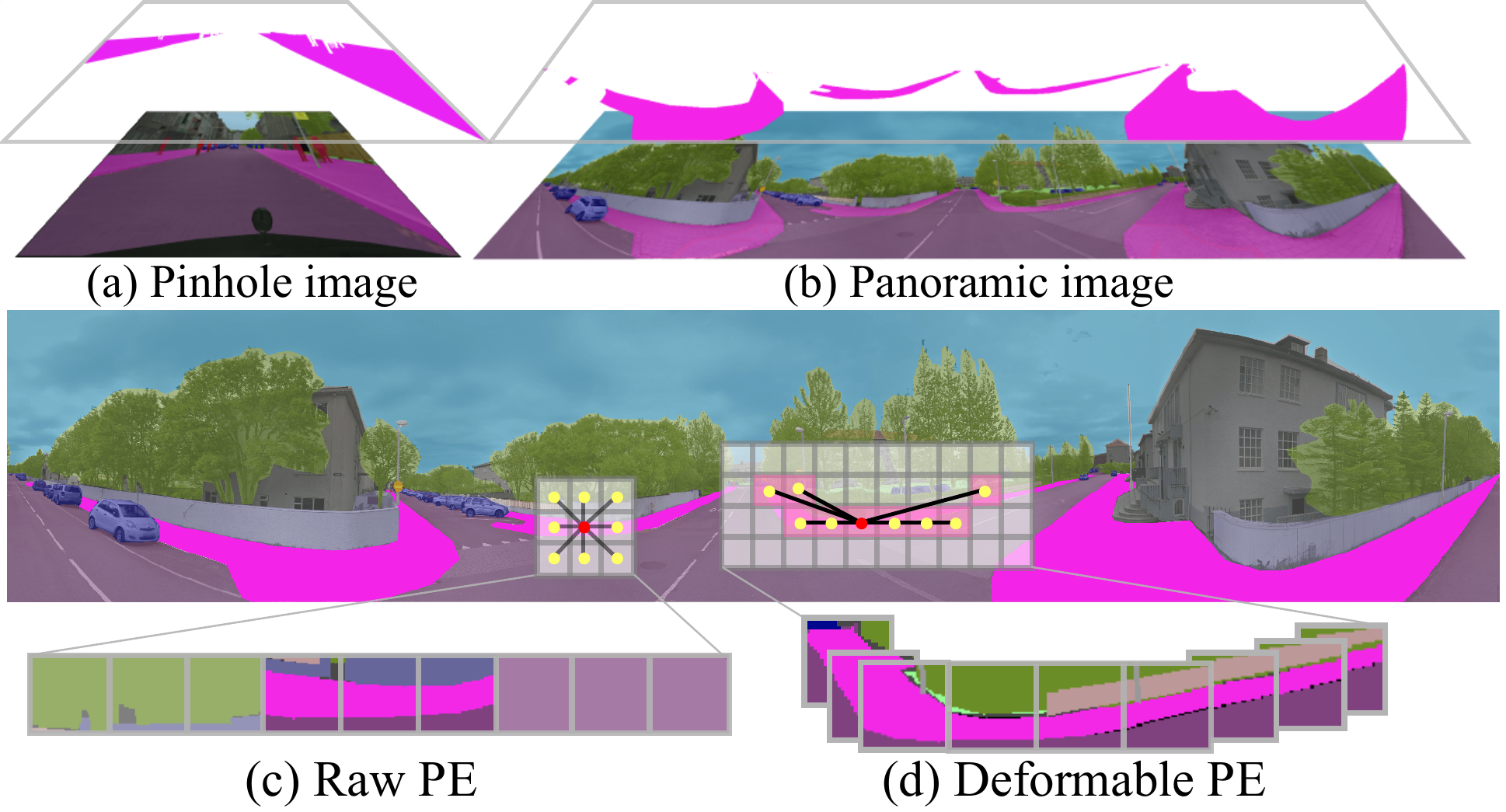}
    \vskip -2ex
	\caption{\small Semantic segmentation of (a) narrow-angle pinhole image and (b) $360^\circ$ panoramic image. Compared to (c) standard Patch Embeddings, our (d) Deformable Patch Embedding partitions $360^\circ$ images while considering distortions, \eg in \emph{sidewalks}. 
	} 
    \label{fig:distortion}
\vskip -4ex
\end{figure}

\section{Introduction}
Panoramic $360^\circ$ cameras have received an increasing amount of attention in fields, such as omnidirectional sensing in automated vehicles~\cite{Garanderie_2018_ECCV,omnirange} and bringing immersive viewing experiences to augmented- and virtual reality displays~\cite{xu2018predicting_head_movement,xu2021spherical}.
Opposed to images captured with pinhole cameras, that occupy narrow Fields of View~(FoV), panoramic images offer omni-range perception, benefiting the detection of road scene objects and indoor scene elements~\cite{Garanderie_2018_ECCV,whats_in_my_room}.
In particular, dense semantic segmentation on panoramic images, facilitates a high-level holistic pixel-wise understanding of surrounding environments~\cite{densepass,dspass}.

Panoramic semantic segmentation is usually performed on 2D panoramas that were transformed using equirectangular projection~\cite{hohonet,omnirange}, which is accompanied by image distortions and object deformations (see Fig.~\ref{fig:distortion}).
Further, in the $360^\circ$ image domain, labeled data is scarce which necessitates model training to be carried out on semantically matching narrow-FoV pinhole datasets.
These two circumstances culminate in a significantly degraded performance on panoramic segmentation
as compared to the pinhole counterpart~\cite{pass} and as such they have to be adequately addressed.
Considering the intricacies of panoramas, convolution variants~\cite{gauge_equivariant,equivariant_networks,distortion_aware} and attention-augmented models~\cite{omnirange} were proposed to mitigate image distortions and enlarge receptive fields of Convolutional Neural Networks (CNNs). However, they remain sub-optimal in handling the severe deformations from pinhole- to panoramic data, and fail in establishing long-range contextual dependencies in the ultra-wide $360^\circ$ images, which prove essential for accurate semantic segmentation~\cite{danet,setr}.

In light of these challenges, we propose a \emph{Transformer for PAnoramic Semantic Segmentation (Trans4PASS)} architecture, and overcome image distortions and object deformations with two novel design choices: Our Deformable Patch Embedding (DPE) is located at the early image sequentialization- and intermediate feature interpretation stages empowering the model to learn characteristic panoramic image distortions and preserve semantics. Secondly, with the Deformable MLP (DMLP) module in the feature parsing stage, we mix patches with learned spatial offsets to enhance global context modeling.

The challenging mismatch between the label-rich pinhole- and the label-scarce panoramic domain can also be addressed by unsupervised domain adaptation (UDA), considering labeled 2D Pinhole images as source- and $360^\circ$ Panoramas as target domain.
Following previous works~\cite{densepass,omnirange}, we refer to this scenario as \textsc{Pin2Pan}.
Taking this view on the learning problem, shows to be a vital ingredient for circumventing the expensive panoramic image annotation process while satisfying the need for large-scale annotated data~\cite{setr} to train robust segmentation transformers.
Unlike common adversarial-learning~\cite{clan} and pseudo-label self-learning~\cite{crst} methods for UDA, we put forward \emph{Mutual Prototypical Adaptation (MPA)}, which generates mutual prototypes for pinhole- and panoramic multi-scale feature embeddings, distilling prototypical knowledge of both domains, which proves advantageous to domain-separate distillation~\cite{yue2021pcs}. On top, we show MPA works with pseudo-labels in a joint manner and provides a complementary alignment incentive in the feature space.

To verify the capability for generalization to diverse scenarios of our solution, we evaluate Trans4PASS on both indoor- and outdoor panoramic-view datasets, \ie, Stanford2D3D~\cite{stanford2d3d} and DensePASS~\cite{densepass} benchmarks.
On DensePASS, it outperforms the previous best result~\cite{p2pda_trans} by ${>}10.0\%$ in mIoU. Our solution achieves top performance among unsupervised methods on Stanford2D3D and even ranks higher than many competing supervised methods.

In summary, we deliver the following contributions:
\begin{compactitem}
\item[(1)] We consider panoramic deformations in our distortion-aware Transformer for Panoramic Semantic Segmentation~(\emph{Trans4PASS}) with \emph{deformable patch embedding-} and \emph{deformable MLP} modules.
\item[(2)] We present \emph{Mutual Prototypical Adaptation} to transfer models via distilling dual-domain prototypical knowledge, boosting performance by coupling it with pseudo-labels in feature- and output space. 
\item[(3)] Our framework for transferring models from \textsc{Pin2Pan} yields excellent results on two competitive benchmarks: On Stanford2D3D we circumvent using $1,400$ expensive panorama labels while achieving comparable results and on DensePASS we boost state-of-the-art performance by an absolute $14.39\%$ in mIoU.
\end{compactitem}


\section{Related Work}
\noindent\textbf{Semantic- and panoramic segmentation.}
Dense semantic segmentation is experiencing steep progress since FCN~\cite{fcn} addressed it end-to-end. Following works built upon FCN to improve performance by enlarging receptive fields~\cite{hou2020strip,pspnet} and refining context priors~\cite{jin2021mining,context_prior}.
Driven by non-local blocks~\cite{nonlocal}, self-attention~\cite{attention} is integrated to learn long-range dependencies~\cite{danet,ccnet} within FCNs.
Currently, architectures which replace convolutional- with transformer-based backbones~\cite{vit,touvron2021deit} emerge.
Then, image perception is viewed from the lens of sequence-to-sequence learning with dense prediction transformers~\cite{swin,yuan2021hrformer} and semantic segmentation transformers~\cite{segmenter,setr}.
Recently, MLP-like architectures~\cite{asmlp,liu2021gmlp,mlp_mixer} which alternate spatial- and channel mixing sparked interest for recognition tasks.
Most methods are designed for narrow-FoV images and often have large accuracy drops in the 360$^\circ$ domain.
In this work, we address panoramic segmentation, with a novel Transformer architecture which considers a broad FoV already in its design and handles the panorama-specific semantic distribution via MLP-based mixing.

By capturing wide-FoV scenes, panoramic images can serve as starting point for a more holistic scene understanding.
Outdoor panorama segmentation works rely on fisheye cameras~\cite{restricted,omniscape,universal,woodscape} or panoramic images~\cite{pps,orhan2021semantic_outdoor_panoramic,synthetic,wildpass} for seamless $360^\circ$ parsing.
Indoor methods on the other hand focus on either distortion-mitigated representations~\cite{spherical_unstructured_grids,spherephd,equivariant_networks} or multi-task schemes~\cite{pano_sfmlearner,hohonet,zhang2021deeppanocontext}. 
Most of these works assume that labeled images are available in the target panorama domain.
We cut this requirement for labeled target data and circumvent the prohibitively expensive annotation process of determining pixel-wise semantics in complex real-world surroundings.
Therefore, unlike previous works, we look through the lens of unsupervised transfer learning and introduce a pinhole- to panorama (\textsc{Pin2Pan}) adaptation method to profit from rich, readily available annotated pinhole datasets.
In experiments, our panoramic segmentation transformer architecture generalizes to both indoor and outdoor scenes. 

\noindent\textbf{Unsupervised domain adaptation.}
Domain adaptation has been thoroughly investigated to enhance model generalization to unseen domains, with two predominant paradigms based either on self-training~\cite{cheng2021dual_path,guo2021metacorrection,curriculum_da} or adversarial learning~\cite{all_about_structure,cycada,adaptsegnet}. Self-training methods generally create pseudo-labels to gradually adapt through iterative improvement~\cite{pycda}, whereas adversarial solutions leverage the idea of GANs~\cite{gan} to perform image translation~\cite{cycada,li2019bidirectional}, or enforce alignment in layout matching~\cite{contextual_relation_consistent_da,content_consistent_matching_da} and feature agreement~\cite{clan,densepass}.
Further adaptation flavors, consider uncertainty reduction~\cite{fleuret2021uncertainty_reduction,rectifying}, model ensembling~\cite{maximum_squares_loss,fda}, category-level alignment~\cite{liu2021bapa,ma2021coarse_fine}, or adversarial entropy minimization~\cite{intra_da,advent}.
Relevant to our work, PIT~\cite{gu2021pit} addresses the camera gap with FoV-based adaptation, whereas P2PDA~\cite{densepass} first tackles \textsc{Pin2Pan} transfer by learning attention correspondences.
Aside from distortion-adaptive architecture design, we revisit \textsc{Pin2Pan} segmentation from a feature prototype adaptation-based perspective where we distill panoramic knowledge through class-wise prototypes.
Different from methods using individual prototypes for source and target domains~\cite{yue2021pcs,proda}, we present mutual prototypical adaptation, which jointly exploits source and target feature embeddings to boost transfer beyond the FoV.

\section{Methodology}
Here, we put forward our panoramic semantic segmentation framework. In Sec.~\ref{sec:trans4pass}, we introduce the \emph{Trans4PASS} architecture for capturing distortion-aware features and long-range dependencies, with detailed descriptions of \emph{deformable patch embeddings} and the \emph{deformable MLP} module in Sec.~\ref{sec:dpe} and~\ref{sec:dmlp}. Finally, we outline our domain adaptation method using mutual prototype features in Sec.~\ref{sec:mpa}.

\begin{figure*}[!t]
	\centering
    \includegraphics[width=1.0\textwidth]{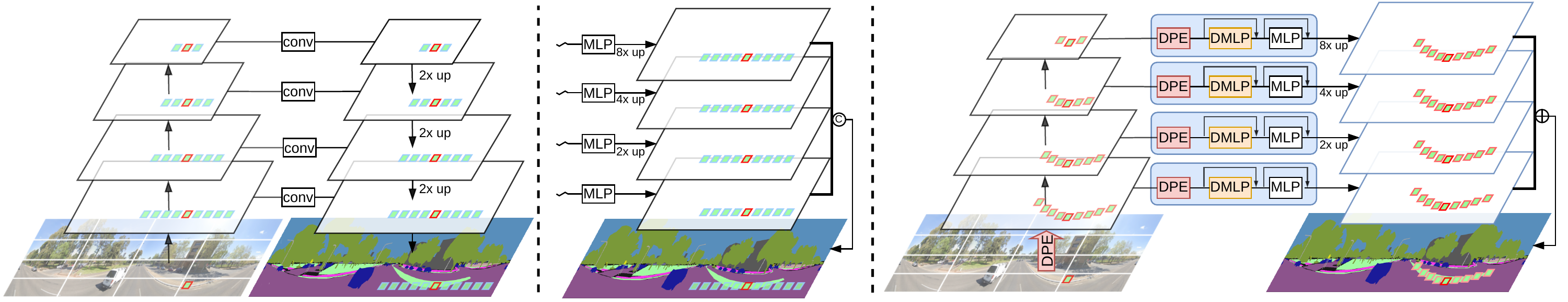}
    \begin{minipage}[t]{.3\textwidth}
    \centering
    \vskip -2ex
    \subcaption{Transformer with FPN-like decoder}\label{fig:fpn}
    \end{minipage}%
    \begin{minipage}[t]{.3\textwidth}
    \centering
    \vskip -2ex
    \subcaption{Transformer with vanilla-MLP}\label{fig:vanilla-mlp}
    \end{minipage}%
    \begin{minipage}[t]{.4\textwidth}
    \centering
    \vskip -2ex
    \subcaption{Trans4PASS with DPE and DMLP}\label{fig:trans4pass}
    \end{minipage}%
    \vskip -3ex
	\caption{\small \textbf{Comparison of segmentation transformers.}
	Transformers (a) borrow a FPN-like decoder~\cite{setr} from CNN counterparts or (b) adopt a vanilla-MLP decoder~\cite{segformer} for feature fusion, which lacks patch mixing.
	(c) \emph{Trans4PASS} integrates Deformable Patch Embeddings~(DPE) and the Deformable MLP~(DMLP) module for capabilities to handle distortions (see warped \emph{terrain}) and mix patches.
	} 
	\label{fig:decoder_structures}
\vskip -3ex
\end{figure*}

\subsection{Trans4PASS Architecture}
\label{sec:trans4pass}
To investigate the transformer model on panoramic semantic segmentation, we create two versions of \emph{Trans4PASS} models (T: Tiny and S: Small). We build both with four stages, where for the tiny model, each stage encompasses $2$ layers, for the small version the stages have $3$, $4$, $6$, and $3$ layers.
As shown in Fig.~\ref{fig:decoder_structures}, the pyramidal stages are inspired by recent transformers~\cite{pvt,segformer}, which reduce the feature scales in deeper layers. Given an input image with $H{\times}W{\times}3$, Trans4PASS makes use of a Patch Embedding~(PE) module~\cite{segformer} to split the image into patches.
To deal with the severe distortions in panoramas, a special \emph{Deformable Patch Embedding~(DPE)} module is proposed and applied in the encoder and decoder (Fig.~\ref{fig:trans4pass}).
In the encoder, each feature map $\boldsymbol{f}_l{\in}\{\boldsymbol{f}_1,\boldsymbol{f}_2,\boldsymbol{f}_3,\boldsymbol{f}_4\}$ in the $l^{th}$ stage is down-sampled by the $l^{th}$ stride ${\in}\{4, 8, 16, 32\}$. The channel dimensions $C_l{\in}\{64,128,320,512\}$ grow successively. 
Different from the FPN-like decoder~\cite{setr} and vanilla-MLP based decoder~\cite{segformer} in Fig.~\ref{fig:decoder_structures},
we propose the \emph{Deformable MLP (DMLP)} decoder structure, which mixes feature patches extracted via DPE.
Given the extracted feature hierarchy in multiple scales from the encoder, four deformable decoder layers process the feature hierarchy into a consistent shape of $\frac{H}{4}{\times}\frac{W}{4}{\times}C_{emb}$, where we set the number of resulting embedding channels $C_{emb}{=}128$.
An ensuing linear layer transforms the $128$ channel output to contain the number of semantic classes of the respective task.

\subsection{Deformable Patch Embedding}
\label{sec:dpe}
Spherical topological images captured by $360^\circ$ cameras occupy a polar coordinate system with $\theta{\in}[0, 2\pi)$ and $\phi {\in} [0, \pi]$. To represent it in 2D space, the spherical data is usually converted into a panoramic format in euclidean-like space through the equirectangular projection.
This process leads to severe shape distortions in the projected panoramic image, as seen in Fig.~\ref{fig:distortion}.
Therefore, a common PE module with fixed sampling positions does not respect these shape distortions of objects and the overall scene. 
Inspired by deformable convolution~\cite{dai2017deformable} and overlapping PE~\cite{segformer}, we propose \emph{Deformable Patch Embeddings (DPE)} and employ them on the input to the encoder and the decoder, splitting panoramic images and features.
Given an input image or feature map $\boldsymbol{f}{\in}\mathbb{R}^{H{\times}W{\times}C_{in}}$, a standard PE module~\cite{vit,segformer} splits it into a flattened 2D patch sequence $\boldsymbol{z}{\in}\mathbb{R}^{(\frac{HW}{s^2}){\times}(s^2 \cdot C_{in})}$, where $\frac{HW}{s^2}$ is the number of patches and $s$ is the width and height of each patch.
Each element in this sequence is passed through a linear projection layer transforming it into $C_{out}$ dimensional embeddings.

Consider a single patch in $\boldsymbol{z}$ representing a rectangle of size $s{\times}s$ with $s^2$ positions. We can define a position offset relative to 
a location $(i,j) | i, j{\in}[1,s]$ in the patch as $\boldsymbol{\Delta}_{(i,j)} {\in} \mathbb{N}^2$.  
In standard PE, these offsets are fixed and lie in $\boldsymbol{\Delta}_{(i, j)}{\in}[\lfloor-\frac{s}{2}\rfloor, \lfloor+\frac{s}{2}\rfloor]^2$.
Take \eg a $3{\times}3$ patch, offsets $\boldsymbol{\Delta}_{(i,j)}$ relative to the center will lie in $[-1, 1] {\times} [-1, 1]$.

As we want to process panoramic images, which inherit distortions from the equirectangular projection, we can directly address this degradation in the PE.
To this end, in our \emph{Deformable Patch Embedding (DPE)}, we enable the model to learn a data-dependent offset $\boldsymbol{\Delta}^{DPE}{\in} \mathbb{N}^{H{\times}W{\times}2}$ that can better cope with the spatial connections of objects, as present in distorted patches.
DPE is learnable and predicts relative offsets based on the original input $\boldsymbol{f}$.
The offset $\boldsymbol{\Delta}^{DPE}_{(i,j)}$ is calculated as depicted in Eq.~\eqref{eq:offset}. \begin{equation}\label{eq:offset}
\begin{aligned}
    \boldsymbol{\Delta}^{DPE}_{(i,j)} &= \begin{bmatrix}
           \min(\max(-\frac{H}{r}, g(\boldsymbol{f})_{(i,j)}), \frac{H}{r}) \\
           \min(\max(-\frac{W}{r}, g(\boldsymbol{f})_{(i,j)}), \frac{W}{r})
         \end{bmatrix},
\end{aligned}
\end{equation}
where $g(\cdot)$ is the offset prediction function, which we implement via the deformable convolution operation~\cite{dai2017deformable}. The hyperparameter $r$ puts a constraint onto the offsets and is set as $4$ in our experiments. 
The learned offsets make DPE adaptive and as a result distortion-aware. 

In earlier works, DPT~\cite{chen2021dpt} applies non-overlapping PE with anchor-based offsets at later stages, PS-ViT~\cite{yue2021psvit} uses a progressive sampling module coupled with previous iterations, and Deformable DETR~\cite{deformable_detr} leverages deformable attention to enhance feature maps.
Unlike these previous works, our proposed DPE is designed for pixel-dense prediction tasks and is flexible to replace the raw PE without having to couple previous iterations. 
Intuitively, a model supplied with DPE, can profit from pinhole images and better adapt to distortions in panoramic images by learning to counteract severe deformations in the data.

\subsection{Deformable MLP}
\label{sec:dmlp}

Apart from the specific design of the encoder, the decoder with an adaptive feature parsing capacity is crucial in segmentation transformers~\cite{segformer,zhang2021trans4trans_acvr}. As shown in Fig.~\ref{fig:fpn}, some transformers~\cite{setr} borrow a FPN-like decoder from the CNN counterpart~\cite{lin2017feature}, whose receptive field is limited to the feature resolution in its final stage~\cite{pvt}. SegFormer~\cite{segformer} takes inspiration from Multilayer Perceptron-based (MLP) models~\cite{mlp_mixer} and integrates a vanilla MLP to combine features (Fig.~\ref{fig:vanilla-mlp}), but does not consider potential distortions in the imaging data.
Next, we propose a mechanism to associate self-attention in Transformers and deformation-properties in $360^\circ$ imagery.
Linking both of these enables profiting from long-range dependencies for dense scene parsing and keeping this improvement when processing panoramic scenes.
Achieving this distortion-aware property at manageable computational complexity, we put forward the \emph{Deformable MLP~(DMLP)} module.
Within each stage of the decoder, DMLP mixes patches across the channel dimension, but with a particularly large receptive field, which improves the interpretation of features delivered by the aforementioned DPE.

\begin{figure}[!t]
	\centering
    \includegraphics[width=1.0\columnwidth]{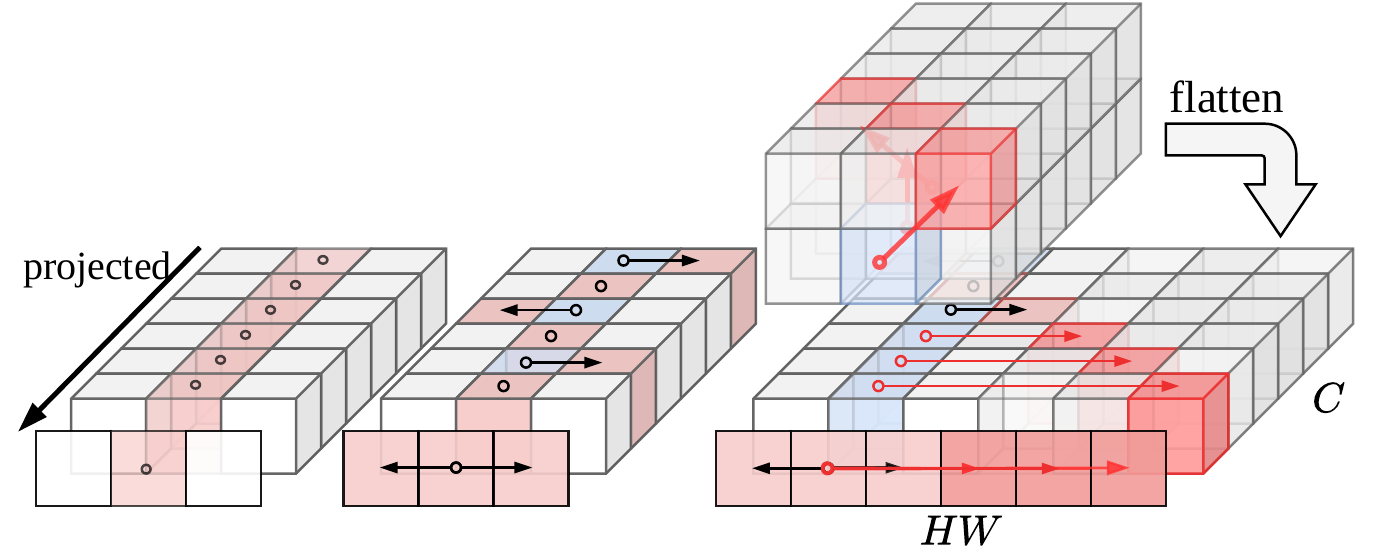}
    \begin{minipage}[t]{.2\columnwidth}
    \centering
    \vskip -3ex
    \subcaption{MLP}\label{fig:MLP}
    \end{minipage}%
    \begin{minipage}[t]{.3\columnwidth}
    \centering
    \vskip -3ex
    \subcaption{CycleMLP}\label{fig:CycleMLP}
    \end{minipage}%
    \begin{minipage}[t]{.5\columnwidth}
    \centering
    \vskip -3ex
    \subcaption{DMLP}\label{fig:OurMLP}
    \end{minipage}%
    \vskip -3ex
	\caption{\small \textbf{Comparison of MLP blocks.} The spatial offsets of DMLP are learned adaptively from the input feature map.} 
	\label{fig:MLPBlocks}
\vskip -3ex
\end{figure}
Fig.~\ref{fig:MLPBlocks} shows the difference in MLP-based modeling: while the vanilla MLP~(see Fig.~\ref{fig:MLP}) performs traditional linear projection without learning any \textcolor{lightred}{spatial context}, CycleMLP (see Fig.~\ref{fig:CycleMLP}) has a limited spatial receptive field by hand-crafted, fixed offsets in mixing patches and their channels.
In Fig.~\ref{fig:OurMLP}, the proposed DMLP generates a learned spatial offset (top) in a wider range and an adaptive manner.
Given the input feature map $\boldsymbol{f}{\in}\mathbb{R}^{H{\times}W{\times}C_{in}}$, the spatial offset $\boldsymbol{\Delta}^{DMLP}_{(i,j,c)}$ is predicted channel-wise as in Eq.~\eqref{eq:offset} and is then flattened as $\boldsymbol{\Delta}^{DMLP}_{(k,c)}$, where $k{\in}{HW}$ and $c{\in}C_{in}$, for mixing the flattened patch features $\boldsymbol{z}{\in}\mathbb{R}^{HW{\times}C_{in}}$, as:
\begin{equation}\label{eq:dmlp}
\begin{aligned}
    \hat{\boldsymbol{z}}_{(k,c)}=\sum_{k=1}^{HW}\sum_{c=1}^{C_{in}}w^{T}_{(k,c)}\cdot{\boldsymbol{z}_{(k+\boldsymbol{\Delta}^{DMLP}_{(k,c)},c)}},
\end{aligned}
\end{equation}
where $w{\in}\mathbb{R}^{C_{in}{\times}C_{out}}$ is the weight matrix of a fully-connected~(FC) layer. 
As shown in Fig~\ref{fig:trans4pass}, the decoder has a similar structure as a MLP-Mixer block~\cite{mlp_mixer}, consisting of DPE, DMLP, and MLP modules. The residual connections are kept. Formally, the four-stage decoder is denoted as:
\begin{equation}\label{eq:parser}
\begin{aligned}
    \hat{\boldsymbol{z}_{l}}&=\textbf{DPE}(C_l, C_{emb})(\boldsymbol{z}_l), \forall_l{\in}\{1,2,3,4\} \\
    \hat{\boldsymbol{z}_{l}}&=\textbf{DMLP}(C_{emb}, C_{emb})(\hat{\boldsymbol{z}_{l}}) + \hat{\boldsymbol{z}_{l}}, \forall_l \\
    \hat{\boldsymbol{z}_{l}}&=\textbf{MLP}(C_{emb}, C_{emb})(\hat{\boldsymbol{z}_{l}}) + \hat{\boldsymbol{z}_{l}}, \forall_l \\
    \hat{\boldsymbol{z}_{l}}&=\textbf{Up}(H/4, W/4)(\hat{\boldsymbol{z}_{l}}), \forall_l \\
    p&=\textbf{LN}(C_{emb}, C_K)(\sum_{l=1}\hat{\boldsymbol{z}_{l}}),
\end{aligned}
\end{equation}
where \textbf{Up}($\cdot$) and \textbf{LN}($\cdot$) refer to the Upsample- and LayerNorm operations, and $p$ is the prediction of $K$ classes.

\subsection{Mutual Prototypical Adaptation}
\label{sec:mpa}
\begin{figure}[!t]
	\centering
    \includegraphics[width=1.0\columnwidth]{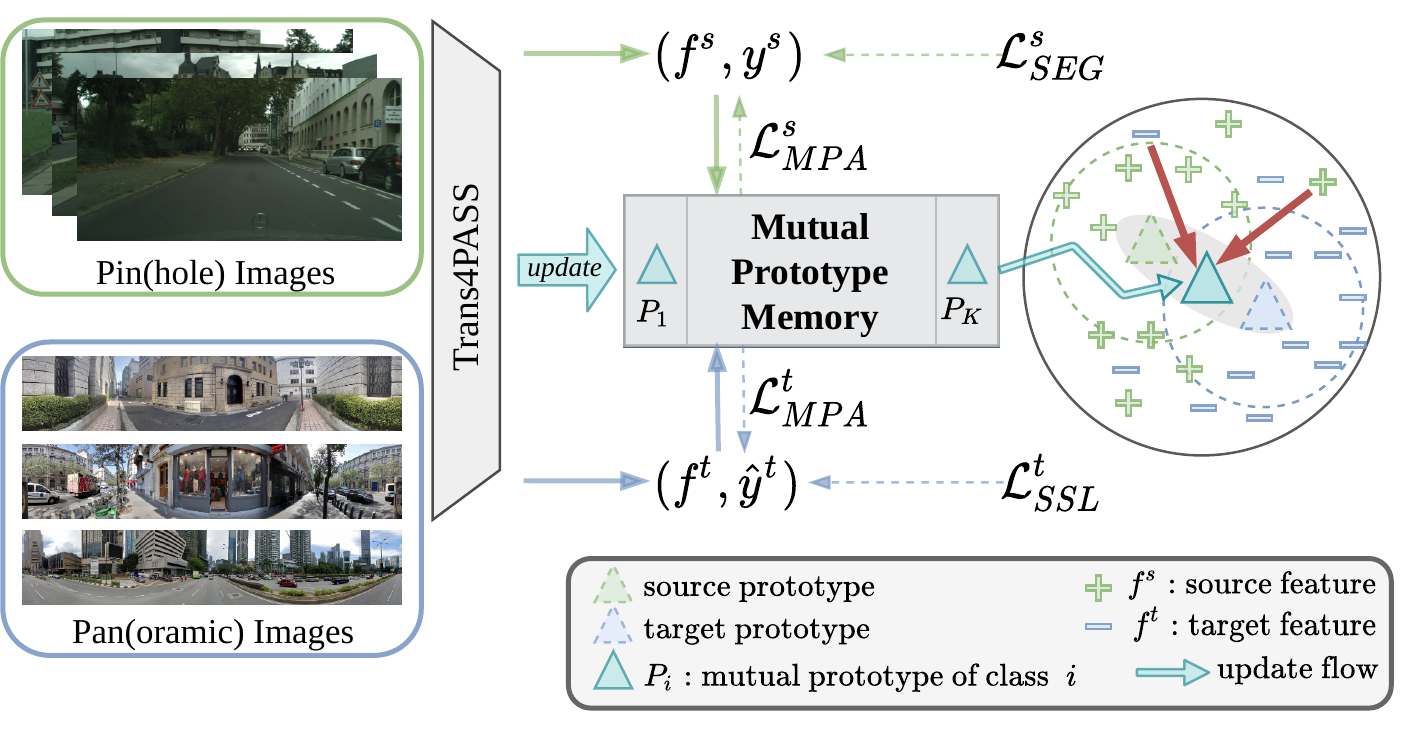}
    \vskip -3ex
	\caption{ \textbf{Diagram of mutual prototypical adaptation.}} 
    \label{fig:mpa}
\vskip -3ex
\end{figure}

Due to the lack of large-scale training data in panoramas, we look into \textsc{Pin2Pan} domain adaptation from a perspective of semantic prototypes~\cite{proda}.
We propose the \emph{Mutual Prototypical Adaptation (MPA)} method to enable distilling knowledge via prototypes which we cultivate through source ground truth labels and target pseudo labels.
Pseudo-labels depend on the few remaining mutual properties from pinhole and panoramic images, \eg, scene distribution at the frontal viewing angle~\cite{hanet,omnirange}.
While the related PCS~\cite{yue2021pcs} performs inter- and intra-domain instance-prototype learning, our mutual prototypes are learned from source- and target feature embeddings $f^s$ and $f^t$, projected to a shared latent space, and stored in a dynamic bank, as shown in Fig.~\ref{fig:mpa}. 
The key differences to PCS lie in that (1) the mutual prototypes are built by joining embeddings from both domains, and (2) our method leverages multi-scale pyramidal features using different input scales in computing the embeddings which yields more robust prototypes.

Given the source (pinhole) dataset with annotated images $\mathcal{D}^s{=}\{(x^s, y^s)|x^s {\in}\mathbb{R}^{H{\times}W{\times}3}, y^s{\in} \{0,1\}^{H{\times}W{\times}K}\}$ and the target (panoramic) dataset $\mathcal{D}^t{=}\{(x^t)|x^t{\in} \mathbb{R}^{H{\times}W{\times}3}\}$ without annotations, the goal of domain adaptation is to learn semantics from the source domain and transfer it to the target domain with $K$ shared classes. The network is trained in $\mathcal{D}^s$ based on the segmentation loss:
\begin{equation}\label{eq:seg}
\begin{aligned}
    \mathcal{L}_{SEG}^s = -\sum_{i,j,k=1}^{H,W, K}y^s_{(i,j,k)}\text{log}(p^s_{(i,j,k)}),
\end{aligned}
\end{equation}
where $p^s_{(i,j,k)}$ indicates the probability of pixel $x^s_{(i,j)}$ predicted as $k$-th class on the source domain.
To generalize the source pre-trained model to the target data, a typical Self-Supervised Learning (SSL) scheme optimizes the model based on the pseudo labels $\hat{y}^t_{(i,j,k)}$ of pixels $x^t_{(i,j)}$ in the target domain: 
\begin{equation}\label{eq:ssl}
\begin{aligned}
    \mathcal{L}_{SSL}^t = -\sum_{i,j,k=1}^{H,W,K}\hat{y}^t_{(i,j,k)}\text{log}(p^t_{(i,j,k)}),
\end{aligned}
\end{equation}
where the pseudo label is given by the most probable class in the model predictions:
$\hat{y}^t_{(i,j,k)}=\mathbbm{1}_{k\doteq\text{arg}\max p^t_{(i,j,:)}}$.
However, training with hard pseudo-labels leaves the model sensitive and fragile against errors in its own prediction and has only a limited positive effect on performance. Therefore, we advocate prototype-based alignment in the feature space, which brings two benefits: (1) it \emph{softens} the hard pseudo-labels by using them in feature space instead of as direct targets and (2) it performs \emph{complementary} alignment of semantic similarities in feature space.

Specifically, given a set with all $n_s$ source- and $n_t$ target feature maps $\boldsymbol{F} {=} \{\boldsymbol{f}^s_{1},\dots,\boldsymbol{f}^s_{n_s}\}{\bigcup}\{\boldsymbol{f}^t_{1},\dots,\boldsymbol{f}^t_{n_t}\}$, with feature maps $\boldsymbol{f}$ fused from four-stage multi-scale features $\boldsymbol{f}{=}\sum_{l=1}^4f_l$.
Each feature map is associated either with its respective source ground-truth label or a target pseudo-label.
To compute the mutual prototype memory $\mathbfcal{M}{=}\{P_1,...,P_K\}$ with prototypes $P_k$ we take the mean of all feature vectors (pixel-embeddings) from all feature maps in $\boldsymbol{F}$ that share the class label $k$.
We initialize $\mathbfcal{M}$ by computing the class-wise mean embeddings through the whole dataset and while training we update the prototype $P_k$ at timestep $t$ online by $P^{t+1}_k{\leftarrow}m{P^{t-1}_k}{+}(1{-}m)P^{t}_k$ with a momentum $m{=0.999}$, where $P^{t}_k$ is the mean pixel-embedding among embeddings that share the class-label $k$ in the current mini-batch.
An overview of this procedure is displayed in Fig.~\ref{fig:mpa}. The mutual prototypical adaptation loss is inspired by the knowledge distillation loss~\cite{chen2020simclr_v2}, which drives the feature embedding $\boldsymbol{f}$ to be aligned with the prototypical feature map $\hat{\boldsymbol{f}}$ which is set up, by stacking the prototypes $P_k {\in} \mathbfcal{M}$ according to the pixel-wise class distribution in either the source label or the pseudo-label. The resulting target $\hat{\boldsymbol{f}}$ has the same shape as $\boldsymbol{f}$. For brevity, only the source domain is displayed in Eq.~\eqref{eq:loss_mpa}, which is similar to the target domain.
\begin{equation}\label{eq:loss_mpa}
\begin{aligned}
    \mathcal{L}_{MPA}^s=&-\lambda{\mathcal{T}^2}\textbf{KL}(\phi(\hat{\boldsymbol{f}}^s/\mathcal{T})||\phi(\boldsymbol{f}^s/\mathcal{T}))\\
    &-(1-\lambda)\textbf{CE}(y^s,\phi(\boldsymbol{f}^s)), \\
\end{aligned}
\end{equation}
where $\textbf{KL}(\cdot)$, $\textbf{CE}(\cdot)$, and $\phi(\cdot)$ are  Kullback–Leibler divergence, Cross-Entropy, and Softmax function, respectively. The temperature $\mathcal{T}$ and hyper-parameter $\lambda$ are $20$ and $0.9$ in our experiments.

The final loss is combined with a weight of $\alpha{=}0.001$ as:
\begin{equation}\label{eq:loss_final}
\begin{aligned}
    \mathcal{L}{=}\mathcal{L}^s_{SEG}{+}\mathcal{L}^t_{SSL}{+}\alpha(\mathcal{L}^{s}_{MPA}{+}\mathcal{L}^{t}_{MPA}).
\end{aligned}
\end{equation}
\vskip -3ex

\section{Experiments}
\subsection{Datasets and Settings}

\noindent\textbf{Indoor pin(hole) dataset.}
Stanford2D3D~\cite{stanford2d3d}~(\textbf{SPin} for short) has $70,496$ pinhole images. The dataset is collected in indoor areas and annotated with $13$ categories. Results are averaged over $3$ official folds, unless otherwise stated. 

\noindent\textbf{Indoor pan(oramic) dataset.}
Stanford2D3D~\cite{stanford2d3d}~(\textbf{SPan} for short) has $1,413$ panoramic images. The images are annotated with the same $13$ categories as its pinhole dataset.

\noindent\textbf{Outdoor pin(hole) dataset.}
Cityscapes~\cite{cityscapes}~(\textbf{CS} for short) dataset comprises $2,979$ and $500$ images for training and validation. Images are annotated with $19$ categories.

\noindent\textbf{Outdoor pan(oramic) dataset.}
DensePASS~\cite{densepass}~(\textbf{DP} for short) collected from cities around the world has $2,000$ images for transfer optimization and $100$ labeled images for testing, annotated with the same $19$ classes as Cityscapes.

\noindent\textbf{Implementation settings.}
We train Trans4PASS models with $4$ 1080Ti GPUs with an initial learning rate of $5e{-}5$, scheduled by the poly strategy with power $0.9$ over $200$ epochs. AdamW~\cite{adam} is the optimizer with epsilon $1e{-}8$, weight decay $1e{-}4$ and batch size is $4$ on each GPU. The image augmentations include random resize with ratio $0.5${--}$2.0$, random horizontal flipping, and random cropping to $512{\times}512$. For outdoor datasets, the resolution is $1080{\times}1080$ and batch size is $1$. When adapting the models from \textsc{Pin2Pan}, the resolution of indoor pinhole and panoramic images are $1080{\times}1080$ and $1024{\times}512$ for training, while the outdoor images are set to $1024{\times}512$ and $2048{\times}400$. The image size of indoor and outdoor validation sets are $2048{\times}1024$ and $2048{\times}400$, respectively. Adaptation models are trained within $10K$ iterations on one GPU.

\begin{table}[!t]
\centering
\footnotesize
\resizebox{\columnwidth}{!}{
\renewcommand{\arraystretch}{0.99}
\begin{tabular}{ll|rrc}
    \toprule
    \textbf{Network} & \textbf{Backbone} & \textbf{CS} & \textbf{DP} & \textbf{mIoU Gaps}
    \\ \midrule[0.5pt]
    SwiftNet~\cite{swiftnet} & ResNet-18 & 75.4 & 25.7 & -49.7 \\ 
    Fast-SCNN~\cite{fastscnn}     & Fast-SCNN	   & 69.1  &  24.6   & -44.5 \\
    ERFNet~\cite{erfnet} & ERFNet    & 72.1  & 16.7 &  -55.4  \\ 
    FANet~\cite{fanet} & ResNet-34 & 71.3 & 26.9 & -44.4 \\
    PSPNet~\cite{pspnet}            & ResNet-50    & 78.6  &  29.5  & -49.1 \\
    OCRNet~\cite{ocrnet}            & HRNetV2p-W18 & 78.6  & 30.8  & -47.8 \\
    DeepLabV3+~\cite{deeplabv3+} & ResNet-101   & 80.9  &  32.5  & -48.4 \\
    DANet~\cite{danet}                 & ResNet-101   & 80.4  &  28.5   & -51.9 \\
    DNL~\cite{dnl}                   & ResNet-101   & 80.4  & 32.1  & -48.3 \\
    Semantic-FPN~\cite{panopticfpn} & ResNet-101 & 75.8 &  28.8  & -47.0 \\
    ResNeSt~\cite{resnest}         & ResNeSt-101  & 79.6  &  28.8   & -50.8 \\
    OCRNet~\cite{ocrnet}            & HRNetV2p-W48 & 80.7  &  32.8  & -47.9 \\
    \midrule
    SETR-Naive~\cite{setr} & Transformer-L & 77.9 & 36.1 & -41.8 \\
    SETR-MLA~\cite{setr} & Transformer-L & 77.2 & 35.6 & -41.6 \\
    SETR-PUP~\cite{setr} & Transformer-L & 79.3 & 35.7 & -43.6 \\
    SegFormer-B1~\cite{segformer} & SegFormer-B1 & 78.5 & 38.5 & -40.0 \\
    SegFormer-B2~\cite{segformer} & SegFormer-B2 & 81.0 & 42.4 & -38.6 \\
    \rowcolor{gray!15} Trans4PASS-T & Trans4PASS-T & 79.1 & 41.5 & -37.6   \\
    \rowcolor{gray!15} Trans4PASS-S & Trans4PASS-S & 81.1 & \textbf{44.8} & -36.3  \\
    \bottomrule
\end{tabular}}
\vskip -2ex
\caption{\small \textbf{Performance gaps} of CNN- and transformer-based models from Cityscapes (\textbf{CS})~$@$~1024${\times}$512 to DensePASS (\textbf{DP}). 
}
\label{table:outdoor_domain_gap}
\vskip -2ex
\end{table}

\begin{table}[!t]
\footnotesize
\centering
\resizebox{\columnwidth}{!}{
\renewcommand{\arraystretch}{0.99}
\begin{tabular}{ll|rrc}
    \toprule
    \textbf{Network} & \textbf{Backbone} & \textbf{SPin} & \textbf{SPan} & \textbf{mIoU Gaps}
    \\ \midrule[0.5pt]
    Fast-SCNN~\cite{fastscnn} & Fast-SCNN & 41.71 & 26.86 & -14.85 \\
    SwiftNet~\cite{swiftnet} & ResNet-18 & 42.28 & 34.95 & -7.87 \\
    DANet~\cite{danet} & ResNet-50 & 43.33 & 37.76 & -5.57 \\
    DANet~\cite{danet} & ResNet-101 & 40.09 & 31.81 & -8.28 \\ 
    \midrule
    Trans4Trans-T~\cite{zhang2021trans4trans_acvr} & PVT-T & 41.28 & 24.45 & -16.83 \\ 
    Trans4Trans-S~\cite{zhang2021trans4trans_acvr} & PVT-S & 44.47 & 23.11 & -21.36 \\
    \rowcolor{gray!15} Trans4PASS-T &  Trans4PASS-T  & 49.05 & 46.08 & -2.97  \\
    \rowcolor{gray!15} Trans4PASS-S &  Trans4PASS-S  & 50.20 & \textbf{48.34} & -1.86  \\
    \bottomrule
\end{tabular}}
\vskip -2ex
\caption{\small \textbf{Performance gaps} from Stanford2D3D-Pinhole (\textbf{SPin}) to Stanford2D3D-Panoramic (\textbf{SPan}) dataset on fold-1.}
\label{table:indoor_domain_gap}
\vskip -3ex
\end{table}

\subsection{\textbf{\textsc{Pin2Pan}} Gaps}

\noindent\textbf{Domain gap in outdoor scenarios.} 
To quantify the \textsc{Pin2Pan} domain gap in outdoor scenarios, we evaluate over $15$ off-the-shelf segmentation models trained on Cityscapes.\footnote[1]{MMSegmentation: https://github.com/open-mmlab/mmsegmentation.}
Table~\ref{table:outdoor_domain_gap} summarizes the results tested on Cityscapes and DensePASS validation sets. Although previous transformers~\cite{segformer,setr} reduce the mIoU gap from ${{\sim}50}\%$ of CNN-based counterparts to ${\sim}40\%$, the \textsc{Pin2Pan} gap remains large. The proposed Trans4PASS architecture has a high performance on pinhole image segmentation and also outperforms other methods on panoramic segmentation with $44.8\%$ mIoU without any adaptation strategy.
It indicates that distortion-aware features and long-range cues maintained in both low and high levels of Transformers as opposed to the context learned in higher-levels of CNNs, are important for wide-FoV panoramic segmentation.

\noindent\textbf{Domain gap in indoor scenarios.}
Table~\ref{table:indoor_domain_gap} shows \textsc{Pin2Pan} domain gaps in indoor scenarios.
As pinhole and panoramic images from Stanford2D3D are captured under the same setting, the \textsc{Pin2Pan} gap is smaller compared to the outdoor scenario. Still, in light of other CNN- and transformer-based methods, the small Trans4PASS version achieves $50.20\%$ and $48.34\%$ mIoU in pinhole- and panoramic image segmentation, yielding the smallest performance drop. 

\begin{table}[!t]
\centering
\resizebox{\columnwidth}{!}{
\setlength\tabcolsep{3.0pt}
\begin{tabular}{@{}l@{}ll@{}|rccl@{}}
    \toprule
    \textbf{Network} & \textbf{Encoder} & \textbf{Decoder} & \scriptsize{\textbf{GFLOPs}} & \textbf{\#P} & \textbf{CS} & \textbf{DP}\\ 
    \midrule     \midrule    
    \multicolumn{7}{@{}p{\columnwidth}@{}}{\textit{\textbf{(1) Compare PEs and MLPs:}}} \\ \midrule
    Trans4PASS & MiT-B1${\textbf{*}}$ & DMLP & 13.11 & 13.10 & 69.48 & 36.50 \\
    Trans4PASS & MiT-B1$\dagger$ & CycleMLP~\cite{cyclemlp} & 9.83 & 13.60 & 73.49 & 40.16 \\
    Trans4PASS & MiT-B1$\dagger$ & ASMLP~\cite{asmlp} & 13.40 & 14.19 & 73.65 & 42.05 \\ 
    \rowcolor{gray!15} Trans4PASS & MiT-B1$\dagger$ & DMLP & 12.02 & 13.93 & 72.49 & 45.89~\gbf{+9.39} \\ \midrule[1pt]
    \multicolumn{7}{@{}p{\columnwidth}@{}}{\textit{\textbf{(2) Compare encoders and decoders: }}} \\ \midrule
    PVT~\cite{pvt} & PVT-T & FPN   & 11.17 & 12.76 & 71.46  & 31.20 \\
    PVT~\cite{pvt} & PVT-T & Vanilla MLP   & 14.56 & 12.84 & 70.60 & 32.85 \\
    PVT~\cite{pvt} & PVT-T & DMLP   &  13.11  &  13.10  &  71.75  & 35.18~\gbf{+3.98} \\ 
    \rowcolor{gray!15} Trans4PASS & PVT-T$\dagger$ & DMLP  & 13.18 & 13.10 & 69.62 & 36.50~\gbf{+5.30} \\  \midrule
    SegFormer~\cite{segformer} & MiT-B1   & Vanilla MLP    & 13.27 & 13.66 & 74.93  & 39.02 \\
    SegFormer~\cite{segformer} & MiT-B1   & FPN    & 9.88 & 13.58 & 73.96 & 41.14 \\
    SegFormer~\cite{segformer}       & MiT-B1   & DMLP    &   11.82    &   13.92 &  73.10  & 45.14~\gbf{+6.12} \\
    \rowcolor{gray!15} Trans4PASS                       & MiT-B1$\dagger$ & DMLP & 12.02 & 13.93 & 72.49 & \textbf{45.89}~\gbf{+6.87} \\ 

    \bottomrule
\end{tabular}}
\vskip -2ex
\caption{\small \textbf{Trans4PASS structural analysis.} 
$\textbf{*}$ and $\dagger$ denote DPT~\cite{chen2021dpt} and our DPE. ``\textbf{\#P}'' is short for \#Parameters in millions. Models are trained on Cityscapes~(\textbf{CS})~$@$~512${\times}$512 and tested on DensePASS~(\textbf{DP})~$@$~2048${\times}$400.}
\label{tab:ablation_structure}
\vskip -3ex
\end{table}

\subsection{Trans4PASS Structural Analysis}
\noindent\textbf{Effect of DPE.}
We compare DPE against DePatch from DPT~\cite{chen2021dpt}. 
While the object-aware offsets and scales in DPT make patches shift around the object, our DPE is flexible to split image patches and is decoupled from object proposals.
As shown in the first group of Table~\ref{tab:ablation_structure}, compared with DPT, our DPE-based Trans4PASS adds $+3.01\%$ and $+9.39\%$ mIoU on Cityscapes and DensePASS, respectively.

\noindent\textbf{Effect of DMLP.}
To ablate the effect of different MLP-like modules embedded in the decoder of Trans4PASS, we substitute DMLP by CycleMLP~\cite{cyclemlp} and ASMLP~\cite{asmlp} modules. DMLP is lighter than ASMLP with fewer GFLOPs, parameters and it is more adaptive as opposed to the fixed offsets in CycleMLP. The first group of Table~\ref{tab:ablation_structure} shows that DMLP outperforms both modules with $3\%$ to $5\%$ in mIoU.

\noindent\textbf{Effect of encoders and decoders.}
With the same encoder as PVT, a DMLP-based decoder brings a $+3.98\%$ improvement compared to the FPN- and MLP-based decoders, as shown in the second group of Table~\ref{tab:ablation_structure}. When our DPE is applied in the early stage of the PVT encoder, further improvements of $+5.30\%$ can be made. Similar improvement results ($+6.12\%$ and $+6.87\%$) are evident in experiments with a SegFormer encoder.
Overall, these results show that DPE and DMLP can be integrated into diverse backbones, significantly improving distortion-adaptability for panoramic scene segmentation.

\begin{table*}[!t]
	\renewcommand\arraystretch{1.0}
	\begin{center}
		\begin{subtable}[ht]{\textwidth}
\setlength{\tabcolsep}{4pt}
    \begin{center}
	\resizebox{\textwidth}{!}{
	\begin{tabular}{ @{}l | c | c c c c c c c c c c c c c c c c c c c@{}}
        \toprule[1pt]
        Method & \rotatebox{90}{mIoU} &  \rotatebox{90}{road} &  \rotatebox{90}{sidewalk} &  \rotatebox{90}{building} & \rotatebox{90}{ wall} &  \rotatebox{90}{fence} &  \rotatebox{90}{pole} & \rotatebox{90}{traffic light} &  \rotatebox{90}{traffic sign}&  \rotatebox{90}{vegetation} &  \rotatebox{90}{terrain} &  \rotatebox{90}{sky} & \rotatebox{90}{person} &  \rotatebox{90}{rider} & \rotatebox{90}{car} &  \rotatebox{90}{truck}& \rotatebox{90}{ bus}& \rotatebox{90}{ train}& \rotatebox{90}{ motorcycle}&  \rotatebox{90}{bicycle}\\
        \toprule[1pt]
        ERFNet~\cite{erfnet} & 16.65 & 63.59 & 18.22 & 47.01 & 9.45 & 12.79 & 17.00 & 8.12 & 6.41 & 34.24 & 10.15 & 18.43 & 4.96 & 2.31 & 46.03 & 3.19 & 0.59 & 0.00 & 8.30 & 5.55 \\
        PASS (ERFNet)~\cite{pass} & 23.66 & 67.84 & 28.75 & 59.69 & 19.96 & 29.41 & 8.26 & 4.54 & 8.07 & 64.96 & 13.75 & 33.50 & 12.87 & 3.17 & 48.26 & 2.17 & 0.82 & 0.29 & 23.76& 19.46 \\
        ECANet (Omni-supervised)~\cite{omnirange} & 43.02 & \textbf{81.60} & 19.46 & 81.00 & \textbf{32.02} & 39.47 & 25.54 & 3.85 & 17.38 & 79.01 & 39.75 & \textbf{94.60} & 46.39 & 12.98 & \textbf{81.96} & 49.25 & 28.29 & 0.00 & 55.36 & 29.47 \\
        \midrule
        CLAN (Adversarial)~\cite{clan} & 31.46 & 65.39 & 21.14 & 69.10 & 17.29 & 25.49 & 11.17 & 3.14 & 7.61 & 71.03 & 28.19 & 55.55 & 18.86 & 2.76 & 71.60 & 26.42 & 17.99 & 59.53 & 9.44 & 15.91\\
        CRST (Self-training)~\cite{crst} & 31.67 & 68.18 & 15.72 & 76.78 & 14.06 & 26.11 & 9.90 & 0.82 & 2.66 & 69.36 & 21.95 & 80.06 & 9.71 & 1.25 & 65.12 & 38.76 & 27.22 & 48.85 & 7.10 & 18.08\\
        P2PDA (Adversarial)~\cite{p2pda_trans} & 41.99 & 70.21 & 30.24 & 78.44 & 26.72 & 28.44 & 14.02 & 11.67 & 5.79 & 68.54 & 38.20 & 85.97 & 28.14 & 0.00 & 70.36 & 60.49 & 38.90 & 77.80 & 39.85 & 24.02\\
        SIM (Self-training)~\cite{wang2020differential} & 44.58 & 68.16 & 32.59 & 80.58 & 25.68 & 31.38 & 23.60 & 19.39 & 14.09 & 72.65 & 26.41 & 87.88 & 41.74 & 16.09 & 73.56 & 47.08 & 42.81 & 56.35 & 47.72 & 39.30\\
        PCS (Self-training)~\cite{yue2021pcs} & 53.83 & 78.10 & \textbf{46.24} & 86.24 & 30.33 & \textbf{45.78} & 34.04 & 22.74 & 13.00 & \textbf{79.98} & 33.07 & 93.44 & 47.69 & 22.53 & 79.20 & 61.59 & 67.09 & 83.26 & 58.68 & 39.80 \\
        \midrule
        USSS (IDD)~\cite{usss} & 26.98 & 68.85 & 5.41 & 67.39 & 15.10 & 21.79 & 13.18 & 0.12 & 7.73 & 70.27 & 8.84 & 85.53 & 22.05 & 1.71 & 58.69 & 16.41 & 12.01 & 0.00 & 23.58 & 13.90 \\        
        USSS (Mapillary)~\cite{usss} & 30.87 & 71.01 & 31.85 & 76.79 & 12.13 & 23.61 & 11.93 & 3.23 & 10.15 & 73.11 & 31.24 & 89.59 & 16.05 & 3.86 & 65.27 & 24.46 & 18.72 & 0.00 & 9.08 & 14.48\\
        Seamless (Mapillary)~\cite{seamless} & 34.14 & 59.26 & 24.48 & 77.35 & 12.82 & 30.91 & 12.63 & 15.89 & 17.73 & 75.61 & 33.30 & 87.30 & 19.69 & 4.59 & 63.94 & 25.81 & 57.16 & 0.00 & 11.59 & 19.04 \\
        SwiftNet (Cityscapes)~\cite{swiftnet} & 25.67 & 50.73 & 32.76 & 70.24 & 12.63 & 24.02 & 18.79 & 7.18 & 4.01 & 64.93 & 23.70 & 84.29 & 14.91 & 0.97 & 43.46 & 8.92 & 0.04 & 4.45 & 12.77 & 8.77 \\
        SwiftNet (Merge3)~\cite{issafe} & 32.04 & 68.31 & 38.59 & 81.48 & 15.65 & 23.91 & 20.74 & 5.95 & 0.00 & 70.64 & 25.09 & 90.93 & 32.66 & 0.00 & 66.91 & 42.30 & 5.97 & 0.07 & 6.85 & 12.66 \\
        \midrule
        Trans4PASS-S (ours) & \textbf{55.25} & 78.39 & 41.62 & \textbf{86.47} & 31.56 & 45.47 & 34.02 & 22.98 & 18.33 & 79.63 & 41.35 & 93.80 & 49.02 & 22.99 & 81.05 & 67.43 & \textbf{69.64} & 86.04 & 60.85 & 39.20\\
        \rowcolor{gray!15} Trans4PASS-S (ours)* & \textbf{56.38} & 79.91 & 42.68 & 86.26 & 30.68 & 42.32 & \textbf{36.61} & \textbf{24.81} & \textbf{19.64} & 78.80 & \textbf{44.73} & 93.84 & \textbf{50.71} & \textbf{24.39} & 81.72 & \textbf{68.86} & 66.18 & \textbf{88.62} & \textbf{63.87} & \textbf{46.62}\\
        \bottomrule[1pt]
        \end{tabular}
    }
    \end{center}
    \vskip-3ex
    \caption{\textbf{Per-class results on DensePASS.} Comparison with state-of-the-art panoramic segmentation \cite{pass,omnirange}, domain adaptation~\cite{clan,wang2020differential,yue2021pcs,p2pda_trans,crst}, and multi-supervision methods~\cite{usss,seamless,issafe}. 
    *~denotes performing multi-scale (MS) evaluation.}
    \label{tab:more}
\end{subtable}

\begin{subtable}[ht]{0.29\textwidth}
	\begin{center}
	\footnotesize
	\setlength{\tabcolsep}{2mm}
	\resizebox{\textwidth}{!}{
    \renewcommand{\arraystretch}{0.9}
    \begin{tabular}{ l | c | c }
    \toprule
    \textbf{Network} & \textbf{Method} &  \textbf{mIoU(\%)} \\
    \toprule
    FANet & P2PDA & 35.67 \\
    DANet & P2PDA & 41.99 \\
    Trans4PASS-T & P2PDA & 51.05 \\
    Trans4PASS-S & P2PDA & 52.91 \\
    \midrule
    Trans4PASS-T & -  & 45.89  \\
    Trans4PASS-T & Warm-up  & 50.56 \\
    Trans4PASS-T & SSL  & 51.86 \\
    Trans4PASS-T & MPA  & 51.93 \\
    Trans4PASS-T & MPA + SSL  & 53.26 \\
    \rowcolor{gray!15} Trans4PASS-T & MPA + SSL + MS  & \textbf{54.72} \\
    \midrule 
    Trans4PASS-S & -  & 48.73 \\
    Trans4PASS-S & Warm-up  & 52.59 \\
    Trans4PASS-S & SSL  & 54.67 \\
    Trans4PASS-S & MPA  & 54.77 \\
    Trans4PASS-S & MPA + SSL  & {55.25} \\
    \rowcolor{gray!15} Trans4PASS-S & MPA + SSL + MS  & \textbf{56.38} \\
    \bottomrule[1pt]
    \end{tabular}
	}
	\captionsetup{font={footnotesize}}
    \vskip-1ex
    \caption{\textbf{Adaptation results on DensePASS.}}
    \label{tab:per_class_densepass}
	\end{center}
\end{subtable}
\hspace{3pt}
\begin{subtable}[ht]{0.31\textwidth}
	\begin{center}
    \scriptsize
	\setlength{\tabcolsep}{2mm}
	\resizebox{0.94\textwidth}{!}{
    \renewcommand{\arraystretch}{}
    \begin{tabular}{ l | c | c }
    \toprule
    \textbf{Network} & \textbf{Method} &  \textbf{mIoU(\%)} \\
    \toprule
    DANet & - & 40.28 \\
    DANet & P2PDA & 42.26 \\
    \midrule
    PVT-Tiny & - & 24.45 \\
    PVT-Tiny & P2PDA & 39.66 \\
    PVT-Small & - & 23.11 \\
    PVT-Small & P2PDA & 43.10 \\
    \midrule
    Trans4PASS-T &-& 46.08 \\
    \rowcolor{gray!15} Trans4PASS-T & MPA & 47.48 \\
    Trans4PASS-S &-& 48.34 \\
    \rowcolor{gray!15} Trans4PASS-S & MPA  & \textbf{52.15} \\
    \midrule    
    DANet & Supervised & 44.15 \\
    \rowcolor{gray!15} Trans4PASS-S & Supervised & \textbf{53.31} \\
    \bottomrule
    \end{tabular}
	}
	\captionsetup{font={footnotesize}}
    \vskip-1ex
	\caption{\textbf{Adaptation results on SPan}~$@$~fold-1.
	}
	\label{tab:per_class_s2d3d-pan}
	\end{center}
\end{subtable}
\hspace{3pt}
\begin{subtable}[ht]{0.31\textwidth}
	\begin{center}
	\setlength{\tabcolsep}{1.0mm}
	\resizebox{\textwidth}{!}{
    \renewcommand{\arraystretch}{0.99}
    \begin{tabular}{ c| l | c | c}
    \toprule[1pt]
    &\textbf{Method} & \textbf{Input} & \textbf{mIoU(\%)} \\ \toprule[1pt]
    {\multirow{10}{*}{\rotatebox[origin=c]{90}{\textit{Supervised}}}} 
    &StdConv~\cite{distortion_aware} &RGB&32.6 \\
    &CubeMap~\cite{distortion_aware} &RGB&33.8 \\
    &DistConv~\cite{distortion_aware} &RGB&34.6 \\
    &UNet~\cite{unet}&RGB-D& 35.9 \\
    &GaugeNet~\cite{gauge_equivariant} &RGB-D &39.4  \\
    &UGSCNN~\cite{spherical_unstructured_grids}& RGB-D  & 38.3 \\
    &HexRUNet~\cite{orientation} &RGB-D & 43.3 \\
    &Tangent~\cite{tangent} (ResNet-101) & RGB  &45.6  \\
    &HoHoNet~\cite{hohonet} (ResNet-101) & RGB  & 52.0  \\
    &Trans4PASS (Small) & RGB & 52.1 \\
    &\cellcolor{gray!15}Trans4PASS (Small+MS) & \cellcolor{gray!15}RGB & \cellcolor{gray!15}\textbf{53.0} \\
    \midrule
    {\multirow{3}{*}{\rotatebox[origin=c]{90}{\textit{UDA}}}} 
    &Trans4PASS (Source only) & RGB & 48.1 \\
    &Trans4PASS (MPA)   & RGB  & 50.8 \\
    &\cellcolor{gray!15}Trans4PASS (MPA+MS)   & \cellcolor{gray!15}RGB  & \cellcolor{gray!15}\textbf{51.2} \\
    \bottomrule[1pt]
    \end{tabular}
	}
	\captionsetup{font={footnotesize}}
    \vskip-1ex
    \caption{\textbf{Comparison on SPan} averaged by 3 folds.}
    \label{tab:supervised_s2d3d_pan}
	\end{center}
\end{subtable}

	\end{center}
	\vskip-4ex
	\caption{\textbf{Comparisons and ablation studies of \textsc{Pin2Pan} domain adaptation} in indoor and outdoor scenarios.}
	\label{tab:pin2pan_in_out}
	\vskip-3ex
\end{table*}

\subsection{\textbf{\textsc{Pin2Pan}} Adaptation}
\noindent\textbf{Ablations in outdoor scenarios.}
To verify the generalization ability of applying Trans4PASS in adaptation methods, FANet and DANet used in P2PDA~\cite{densepass} are replaced by Trans4PASS-T/-S, as visible in Table~\ref{tab:per_class_densepass}.
Trans4PASS brings ${>}10\%$ performance gains due to the captured long-range contexts and distortion-aware features.
Without the advantage of a superior network architecture, MPA achieves $51.93\%$ and $54.77\%$ with Trans4PASS-T and -S models, surpassing $51.05\%$ and $52.91\%$ of P2PDA.
The second and third ablation groups of Table~\ref{tab:per_class_densepass} show how Trans4PASS-T and -S match up against each other. 
Individually, MPA is on par with the SSL-based method.
When combining both, MPA and SSL, Trans4Pass-S obtains new state-of-the-art performance on DensePASS, reaching $55.25\%$ in mIoU and $56.38\%$ with multi-scale evaluation. This verifies that MPA works collaboratively with pseudo labels and provides a complementary feature alignment incentive.

\begin{figure}[!t]
	\centering
    \includegraphics[width=1.0\columnwidth]{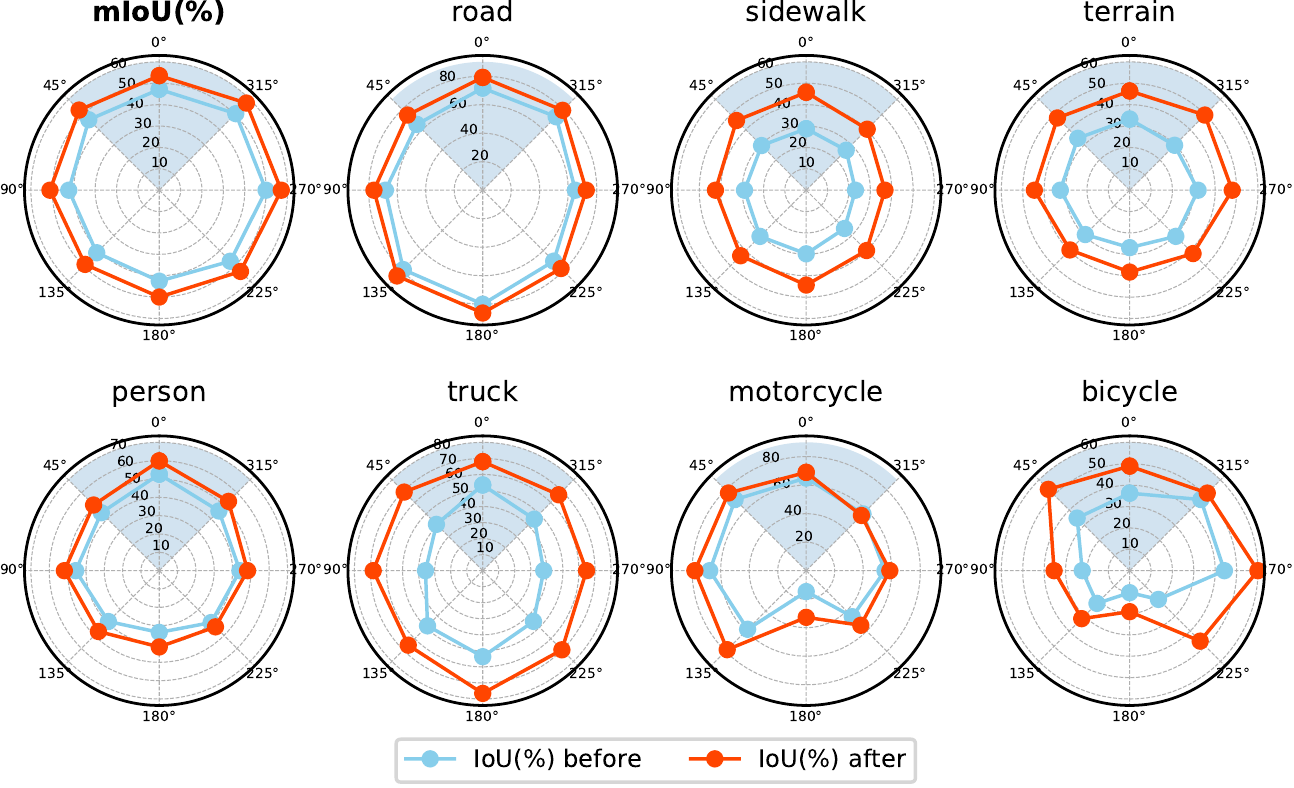}
    \vskip -2ex
	\caption{\textbf{Comparison of omnidirectional segmentation} before and after mutual prototypical adaptation.} 
	\label{fig:polar_densepass}
\vskip -4ex
\end{figure}

\noindent\textbf{Omnidiretional segmentation.}
To showcase the effectiveness of MPA on omnidiretional segmentation, the panoramic image is divided into $8$ directions and evaluated individually.
The polar diagram in Fig.~\ref{fig:polar_densepass} demonstrates that MPA brings uniform improvement to omnidirectional segmentation. Apart from benefiting the stuff classes (\emph{road}, \emph{sidewalk}, and \emph{terrain}), MPA improves the segmentation of object classes, such as \emph{person} and \emph{truck}. Due to the panorama boundary at $180^\circ$, IoUs of \emph{motorcycle} and \emph{bicycle} are impacted, still consistent and large accuracy boosts with MPA in all directions for different classes are observed. 

\noindent\textbf{Comparison with outdoor state-of-the-art methods.}
In Table~\ref{tab:more},
we compare our solution with recent panoramic segmentation~\cite{pass,omnirange} and domain adaptation~\cite{clan,wang2020differential,yue2021pcs,p2pda_trans,crst} methods. Following~\cite{p2pda_trans}, we also involve multi-supervision methods~\cite{usss,seamless,issafe} which require much more data, to broaden the comparison.
MPA-Trans4PASS arrives at the highest mIoU of $56.38\%$, outperforming the previous best P2PDA-SSL on DensePASS by $14.39\%$ and the prototypical method~\cite{yue2021pcs} adapted by Trans4PASS. Trans4PASS obtains top scores on $10$ of $19$ classes.
Notably, our solution shows improvements on challenging categories, \eg, \emph{truck}, \emph{train}, \emph{motorcycle}, and \emph{bicycle}.

\noindent\textbf{Adaptation results in indoor scenarios.}
The experiments in Table~\ref{tab:per_class_s2d3d-pan} are conducted according to the fold-$1$ data splitting~\cite{stanford2d3d} on the Stanford-Panoramic dataset. Our MPA surpasses the previous state-of-the-art P2PDA with DANet and it is even better than the one adapted by a PVT-Small backbone.
Overall, our Trans4PASS-S with MPA achieves the highest mIoU ($52.15\%$), even reaching the level of the fully-supervised Trans4PASS-S ($53.31\%$) which does have access to panoramic image annotations.

\noindent\textbf{Comparison with indoor state-of-the-art methods.}
Before and after adaptation in Table~\ref{tab:supervised_s2d3d_pan}, our Trans4PASS-S model (${\sim}14$M parameters) obtains a high mIoU score ($51.2\%$), even comparable to existing fully-supervised and transfer-learning methods, which are based on ResNet-101 backbones (${\sim}44$M parameters and $52.0\%$ mIoU).

\begin{figure*}[!t]
	\begin{subfigure}[b]{1.0\textwidth}   
		\centering 
		\includegraphics[width=1.0\textwidth]{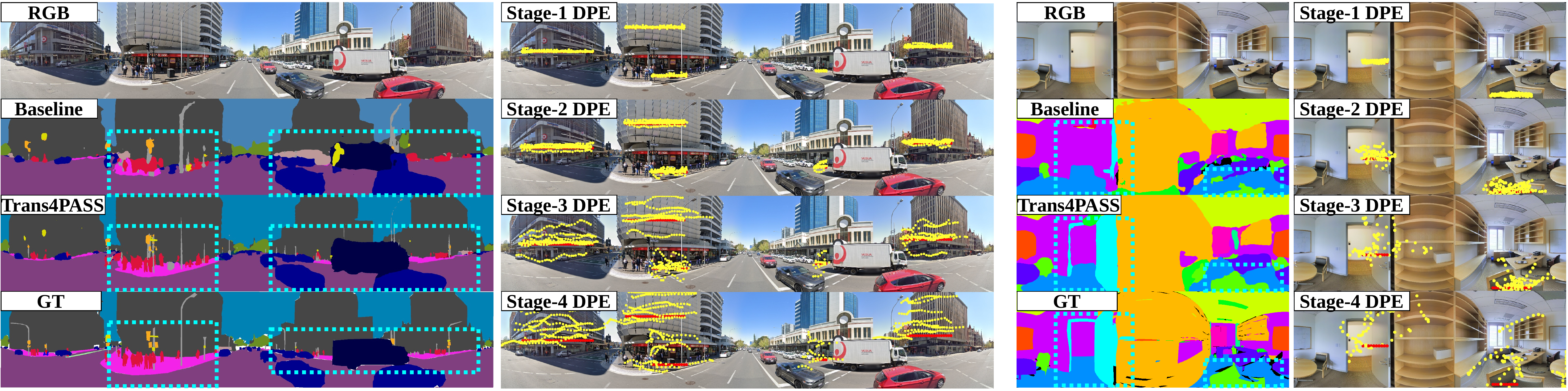}
	    \begin{minipage}[t]{.3\textwidth}
        \centering
        \vskip -2.5ex
        \subcaption{\small Segmentation outdoors}\label{fig:vis_outdoor}
        \end{minipage}%
        \begin{minipage}[t]{.32\textwidth}
        \centering
        \vskip -2.5ex
        \subcaption{\small DPE outdoors }\label{fig:vis_offset_outdoor}
        \end{minipage}%
        \begin{minipage}[t]{.23\textwidth}
        \centering
        \vskip -2.5ex
        \subcaption{\small Segmentation indoors}\label{fig:vis_indoor}
        \end{minipage}%
        \begin{minipage}[t]{.15\textwidth}
        \centering
        \vskip -2.5ex
        \subcaption{\small DPE indoors}\label{fig:vis_offset_indoor}
        \end{minipage}%
		\label{fig:S2D}
	\end{subfigure}
	\vskip -1ex
	\centering
	\begin{subfigure}[b]{1.0\textwidth}   
		\centering 
		\includegraphics[width=1.0\textwidth]{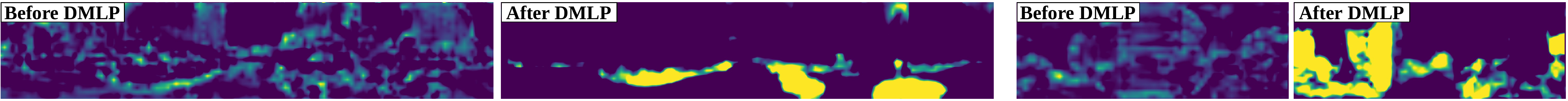}
	    \begin{minipage}[t]{.6\textwidth}
        \centering
        \vskip -2.5ex
        \subcaption{\small DMLP outdoors}\label{fig:vis_dmlp_outdoor}
        \end{minipage}%
	    \begin{minipage}[t]{.4\textwidth}
        \centering
        \vskip -2.5ex
        \subcaption{\small DMLP indoors}\label{fig:vis_dmlp_indoor}
        \end{minipage}%
	\end{subfigure}
    \vskip -3.3ex
	\caption{\small \textbf{Qualitative comparisons, DPE and DMLP visualizations.} (a) and (c) show segmentation comparisons, where the baseline has neither DPE/DMLP nor MPA. The \textcolor{yellow}{$\bullet$} dots in (b) and (d) are sampling points shifted by learned offsets \textit{w.r.t.} the \textcolor{red}{$\bullet$} patch center of DPE (from decoder). (e) and (f) show the $\text{\#}75$ channel maps of stage-$3$ before and after DMLP. 
	Zoom in for better view.} 
	\label{fig:vis}
\vskip -3ex
\end{figure*}

\subsection{Qualitative analysis}
\noindent\textbf{Panoramic semantic segmentation visualizations.}
Fig.~\ref{fig:vis_outdoor} and Fig.~\ref{fig:vis_indoor} demonstrate that Trans4PASS handles the distortion of panoramic images very well as compared to indoor~\cite{pvt} and outdoor~\cite{segformer} baseline models. Especially, the segmentation results for \emph{sidewalks} and \emph{pedestrians} from Trans4PASS have more accurate classifications and boundary distinctions, while the baseline model is confused by the distorted shape and space, due to the lacking capacity to learn long-range contexts and distortion-aware features. In the indoor case of Fig.~\ref{fig:vis_indoor}, the \emph{door} and \emph{chair} categories are barely detected by the baseline model, but our Trans4PASS can output precise segmentation masks on both objects.

\noindent\textbf{DPE and DMLP visualizations.}
Fig.~\ref{fig:vis_offset_outdoor} and Fig.~\ref{fig:vis_offset_indoor} visualize effects of Deformable PE from four stages of Trans4PASS. The red dots denote the centers of a selected patch (size of $s{\times}s$) sequence. Given learned offsets from DPE, $s^{2}$ yellow sampling dots are shifted to semantic-relevant areas in a flexible way,
where each pixel is adaptive to distorted objects and space, like the deformed \emph{building} and \emph{sidewalk} (see Stage-$4$ DPE in Fig.~\ref{fig:vis_offset_outdoor}). Besides, to verify the effect of Deformable MLP, two feature map pairs from the $75^{th}$ channel before and after DMLP are displayed in Fig.~\ref{fig:vis_dmlp_outdoor} and~\ref{fig:vis_dmlp_outdoor}. The feature maps (indoors/outdoors)
after DMLP present semantically recognizable responses, \eg on regions of distorted \emph{sidewalks} or \emph{doors}, as compared to those before the DMLP module.

\section{Conclusion}
To revitalize $360^\circ$ scene understanding, we introduce a universal framework with a Transformer for PAnoramic Semantic Segmentation (Trans4PASS) model and a Mutual Prototypical Adaptation (MPA) method for transferring semantic information from the label-rich pinhole domain to the label-scarce panoramic domain. The Deformable Patch Embedding (DPE) and the Deformable MLP (DMLP) module endow Trans4PASS with distortion awareness. 
The framework elevates state-of-the-art performances on the competitive Stanford2D3D and DensePASS benchmarks.

\noindent\textbf{Limitations.} We note that the accuracy of some classes are still impacted by the partition boundary of panoramas at $180^\circ$. Transferring models between pinhole-, fisheye-, and panoramic domains, fusing modalities, and solving various tasks of $360^\circ$ imagery are opportunities for further research.

\clearpage

{\small
\bibliographystyle{ieee_fullname}
\bibliography{egbib}
}

\newpage
\appendix

\section{Quantitative analysis}

\subsection{Analysis of hyper-parameters}
As the spatial correspondence problem indicated in~\cite{dai2017deformable}, if the deformable convolution is applied to the lower or middle layers, the spatial structures are susceptible to fluctuation~\cite{restricted}. To overcome this problem, we propose the regional restriction of learned offsets to stabilize the training of our early-stage and four-stage Deformable Patch Embedding (DPE) module. Table~\ref{table:effect_r} shows that $r{=}4$ has a better result. Thus, the constraint $r$ applied in the offset prediction module is set as $4$ in our experiments. 

\begin{figure}[t]
	\centering
    \includegraphics[width=0.99\columnwidth]{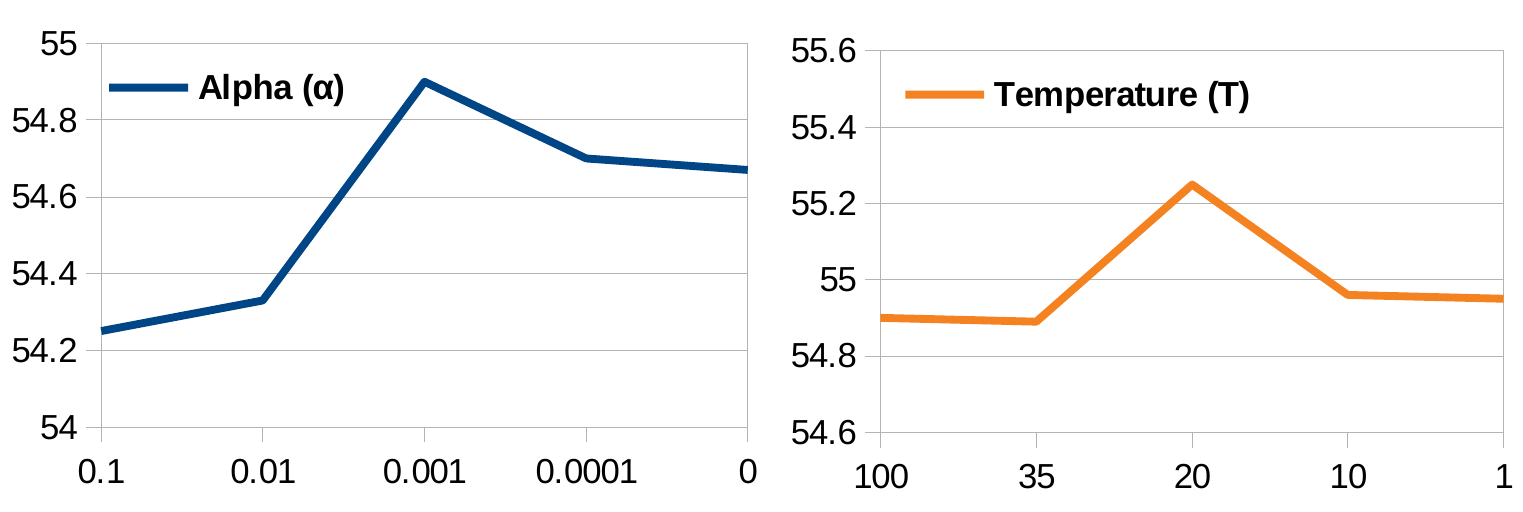}
    \begin{minipage}[t]{.5\columnwidth}
    \centering
    \vskip -3ex
    \subcaption{\small mIoU(\%) {--} $\alpha$}\label{fig:hyper_alpha}
    \end{minipage}%
    \begin{minipage}[t]{.5\columnwidth}
    \centering
    \vskip -3ex
    \subcaption{\small mIoU(\%) {--} $\mathcal{T}$}\label{fig:hyper_temperature}
    \end{minipage}%
    \vskip -2ex
	\caption{ \small \textbf{Analysis of hyper-parameters}. The performance (mIoU) is evaluated in the outdoor target dataset~(DensePASS). } 
    \vskip -0ex
	\label{fig:hyper_params}
\end{figure}

To investigate the effect of various hyper-parameters in the proposed Trans4PASS framework, we analyze the weight $\alpha$ and the temperature $\mathcal{T}$ as shown in Fig.~\ref{fig:hyper_alpha} and Fig.~\ref{fig:hyper_temperature}. The weight $\alpha$ is used to combine the \emph{Mutual Prototypical Adaptation (MPA)} loss and the source- and target segmentation losses. As $\alpha$ decreases from $0.1$ to $0$, we set the temperature $\mathcal{T}{=}35$ in the MPA loss and evaluate the mIoU($\%$) results on the target~(DensePASS~\cite{densepass}) dataset. If $\alpha{=}0$, the final loss is equivalent to that of the SSL-based method, \ie, the MPA loss is excluded. When $\alpha{=}0.001$ for combining both, MPA and SSL, Trans4PASS obtains a better performance. 

Apart from the combination weight $\alpha$, we further investigate the effect of the temperature $\mathcal{T}$, which is used in the MPA loss. As shown in Fig.~\ref{fig:hyper_temperature}, the performance is not sensitive to the distillation temperature, which illustrates the robustness of our MPA method. Nevertheless, we found that MPA performs better when the temperature is lower, so $\mathcal{T}{=}20$ is set as the default setting in our experiments.

\subsection{Computational complexity}
We reported the complexity of Deformable Patch Embedding~(DPE) and Deformable MLP~(DMLP) and compared with other methods on DensePASS in Table~\ref{table:complex_dpe_dmlp}. The results indicate that our methods have significant improvement with the same order of complexity.

\subsection{Detailed results in outdoor scenarios}
Table~\ref{tab_sup:per_class_densepass} shows the per-class IoU results on DensePASS dataset. The first group of experiments is conducted to compare the performance of different backbones in P2PDA~\cite{p2pda_trans} method. Additionally, the adaptation process of the original FANet~\cite{fanet} and DANet~\cite{danet} are shown in more detail, \ie, the performance of the source-only model and that without using the SSL-based method are included. The experiments in the second and third groups are based on Trans4PASS-T and -S model, respectively. As shown in the third group, Trans4PASS-S obtains new state-off-the-art performance in mean IoU (56.38\%). In addition, it achieves top scores on $7$ out of $19$ classes in per-class IoU, including \emph{pole}, \emph{traffic light}, \emph{person}, \emph{car}, \emph{truck}, \emph{motorcycle}, and \emph{bicycle}.

\begin{table}[!t]
\footnotesize
\centering
\resizebox{\columnwidth}{!}{
\begin{tabular}{r|rrrrr}
    \toprule
     & None & r=1 & r=2 & r=4 & r=8 \\ \midrule[0.5pt]
     mIoU(\%) & 45.74 & 44.51& 45.59& \textbf{45.89}& 45.57\\
    \bottomrule
\end{tabular}}
\vskip-1ex
\caption{\small \textbf{Effect of regional restriction ($r$)} on DensePASS.}
\label{table:effect_r}
\vskip-1ex
\end{table}

\begin{table}[!t]
\footnotesize
\centering
\resizebox{\columnwidth}{!}{
\begin{tabular}{@{}r|r>{\columncolor[gray]{0.9}}r|rrr>{\columncolor[gray]{0.9}}r}
    \toprule
     & \textbf{PE}[\textcolor{blue}{79}] & \textbf{DPT}[\textcolor{blue}{8}] & \textbf{DPE} & \textbf{CycleMLP}[\textcolor{blue}{6}] & \textbf{ASMLP}[\textcolor{blue}{43}] & \textbf{DMLP} \\ \midrule[0.5pt]
     GFLOPs & 0.16 & 0.36 & 7.65 & 1.25 & 4.83 & 3.45 \\
     \#Params(M) & 0.01 & 0.02 & 2.90 & 0.45 & 1.04 & 0.79 \\
     mIoU(\%) & 45.14  & \textbf{45.89} & 36.50 & 40.16 & 42.05 & \textbf{45.89} \\
    \bottomrule
\end{tabular}}
\vskip-1ex
\caption{\small \textbf{Computational complexity} of DPE and DMLP. GFLOPs are calculated~@$512{\times}512$.}
\label{table:complex_dpe_dmlp}
\vskip-3ex
\end{table}

\subsection{Detailed results in indoor scenarios}
Apart from the detailed results in outdoor scenarios, per-class results on the outdoor Stanford2D3D-Panoramic dataset~\cite{stanford2d3d} are shown in Table~\ref{tab_sup:per_class_s2d3d_pan}. The experiments are conducted on the fold-$1$ dataset setting of Stanford2D3D~\cite{stanford2d3d}. Our proposed framework with the Trans4PASS-S backbone and the MPA method obtains the best performance in the domain adaptation setting, reaching $52.15\%$ in mean IoU. It also achieves best IoU scores on $7$ out of $13$ classes in the indoor scenario, especially on the \emph{ceiling}, \emph{column}, and \emph{door} categories. In the supervised learning setting, Trans4PASS-S surpasses the CNN-based DANet by a large margin, achieving a score of $53.31\%$ in mean IoU. Besides, its performance in per-class IoU is better than DANet in almost all categories, which lacks the capacity to learn long-range contexts and distortion-aware features in panoramas.

The comparison of segmentation performance with state-of-the-art methods on Stanford2D3D-Panoramic dataset is shown in Table~\ref{tab_sup:supervised_s2d3d_pan}. Since the results of these experiments are based on the average of all $3$ data-splitting settings, we show the results of each individual split setting and its per-class IoU in detail (in \grow{gray}). The small version of the Trans4PASS backbone is used in this experiment. Compared with the previous best fully-supervised method equipped with ResNet-101, Trans4PASS-S has much fewer parameters and is an order of magnitude smaller than ResNet-101. Still, our method obtains the new state-of-the-art performance on Stanford2D3D-Panoramic dataset, reaching $53.0\%$ in mean IoU. Within all $13$ classes, Trans4PASS obtains a total of $8$ best per-class IoUs. In the setting of unsupervised domain adaptation (UDA), our proposed method achieves ${+}2.7\%$ in the average of three folds, and ${+}3.1\%$ when using multi-scale evaluation. It obtains best per-class scores on $9$ out of $13$ categories.

\section{Qualitative analysis}
\subsection{More visualizations in indoor scenarios}
\begin{figure}[!h]
	\begin{subfigure}{\columnwidth}
    \includegraphics[width=0.99\columnwidth]{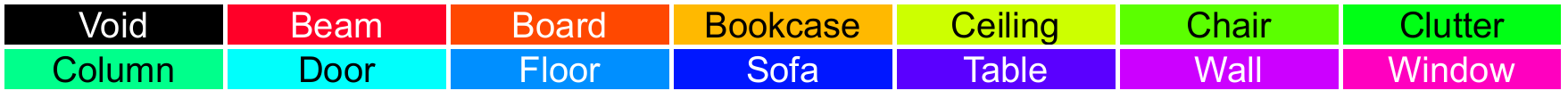}
    \centering \vskip -3ex
    \end{subfigure}
	\centering
    \includegraphics[width=0.99\columnwidth]{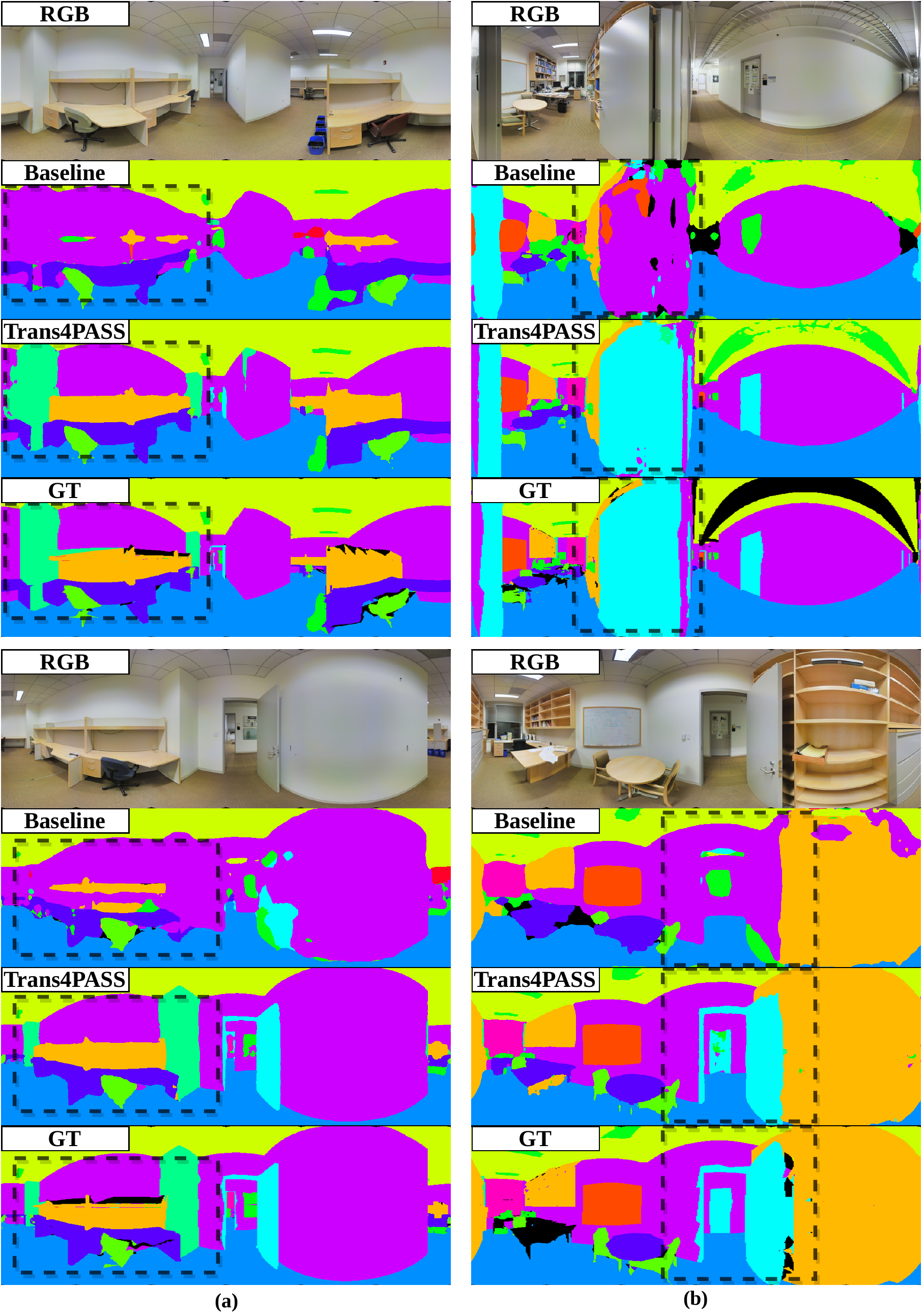}
    \vskip -2ex
	\caption{\textbf{Qualitative comparisons} in indoor scenarios.} 
	\label{fig:supple_vis_indoor}
	\vskip -2ex
\end{figure}

Similar to the visualization in outdoor scenarios, more qualitative comparisons between the baseline and the proposed Trans4PASS are displayed in Fig.~\ref{fig:supple_vis_indoor}, which are from the evaluation set of Stanford2D3D-Panoramic~\cite{stanford2d3d} in the fold-$1$ setting. In Fig.~\ref{fig:supple_vis_indoor}(a), Trans4PASS can produce higher quality segmentation results in those categories highlighted by the black dashed rectangles, such as \emph{column} and \emph{bookcase} categories, while the baseline model can hardly identify these severely deformed objects. In Fig.~\ref{fig:supple_vis_indoor}(b), the \emph{doors} are incorrectly segmented as part of the \emph{wall} by the baseline model, and the correct segmentation results can be generated by our Trans4PASS model.

\begin{table*}[!t]
\renewcommand\arraystretch{0.99}
  \footnotesize
  \setlength{\tabcolsep}{4pt}
  \begin{center}
  \resizebox{\textwidth}{!}{
    \begin{tabular}{ l | c | c | c c c c c c c c c c c c c c c c c c c}
    \toprule[1pt]
    \textbf{Network} & \textbf{Method} &  \textbf{\rotatebox{90}{mIoU}} &  \rotatebox{90}{road} &  \rotatebox{90}{sidewalk} &  \rotatebox{90}{building} & \rotatebox{90}{ wall} &  \rotatebox{90}{fence} &  \rotatebox{90}{pole} & \rotatebox{90}{traffic light} &  \rotatebox{90}{traffic sign}&  \rotatebox{90}{vegetation} &  \rotatebox{90}{terrain} &  \rotatebox{90}{sky} & \rotatebox{90}{person} &  \rotatebox{90}{rider} & \rotatebox{90}{car} &  \rotatebox{90}{truck}& \rotatebox{90}{ bus}& \rotatebox{90}{ train}& \rotatebox{90}{ motorcycle}&  \rotatebox{90}{bicycle}\\
    \toprule
    
    FANet & - & 26.90 &  62.98 & 10.64 & 72.41 & 7.80 & 20.74 & 11.77 & 6.85 & 3.75 & 68.11 & 21.56 & 87.00 & 23.73 & 5.33 & 49.61 & 10.65 & 0.54 & 16.76 & 24.15 & 6.62 \\
    FANet & P2PDA & 33.52 & 57.16 & 25.66 & 78.43 & 16.02 & 26.88 & 12.76 & 2.30 & 7.34 & 68.73 & 26.92 & 87.45 & 36.51 & 1.20 & 62.83 & 20.16 & 0.00 & 68.46 & 17.86 & 20.19 \\
    FANet & P2PDA + SSL & 35.67 & 58.08 & 28.75 & 78.19 & 16.47 & 26.86 & 13.78 & 4.76 & 7.62 & 69.01 & 34.58 & 87.51 & 36.12 & 0.90 & 64.06 & 27.50 & 0.00 & 84.99 & 18.13 & 20.35 \\
    DANet & - & 28.50 & 70.68 & 8.30 & 75.80 & 9.49 & 21.64 & 15.91 & 5.85 & 9.26 & 71.08 & 31.50 & 85.13 & 6.55 & 1.68 & 55.48 & 24.91 & 30.22 & 0.52 & 0.53 & 17.00 \\
    DANet & P2PDA & 40.52 & 62.90 & 25.58 & 76.62 & 24.45 & 30.37 & 14.45 & 16.75 & 9.96 & 67.87 & 19.70 & 82.04 & 34.18 & 22.95 & 56.99 & 54.27 & 44.15 & 47.75 & 46.98 & 31.86  \\
    DANet & P2PDA + SSL & 41.99 & 70.21 & 30.24 & 78.44 & 26.72 & 28.44 & 14.02 & 11.67 & 5.79 & 68.54 & 38.20 & 85.97 & 28.14 & 0.00 & 70.36 & 60.49 & 38.90 & 77.80 & 39.85 & 24.02  \\
    Trans4PASS-T & -  & 45.89  & 72.42 & 32.53 & 84.43 & 20.13 & 35.20 & 24.45 & 15.37 & 12.59 & 78.85 & 31.65 & 90.87 & 42.42 & 14.12 & 74.07 & 39.66 & 35.45 & 90.32 & 50.31 & 26.95  \\
    Trans4PASS-T & P2PDA & 51.05 & 74.82 & 36.53 & 85.93 & 30.23 & 34.83 & 33.70 & 20.36 & 20.40 & 77.43 & 34.87 & 93.65 & 46.01 & 20.89 & 76.85 & 58.19 & 51.20 & 82.19 & 56.84 & 35.09 \\
    Trans4PASS-S & -  & 48.73 & 70.28 & 25.52 & 84.98 & 29.10 & 39.00 & 29.05 & 17.77 & 13.21 & 78.26 & 29.89 & 91.00 & 42.16 & 13.43 & 78.26 & 47.25 & 63.82 & 78.06 & 60.31 & 34.38 \\
    Trans4PASS-S & P2PDA & 52.91 & 76.29 & 41.02 & 86.86 & \textbf{31.96} & 42.15 & 35.15 & 20.98 & 19.49 & 79.44 & 29.26 & 93.64 & 49.62 & 17.47 & 78.77 & 62.80 & 66.38 & 77.98 & 59.23 & 36.73 \\
    \midrule

    Trans4PASS-T & -  & 45.89  & 72.42 & 32.53 & 84.43 & 20.13 & 35.20 & 24.45 & 15.37 & 12.59 & 78.85 & 31.65 & 90.87 & 42.42 & 14.12 & 74.07 & 39.66 & 35.45 & \textbf{90.32} & 50.31 & 26.95  \\
    Trans4PASS-T & Warm-up  & 50.56 & 76.54 & 38.94 & 84.99 & 27.1 & 33.61 & 30.75 & 18.75 & 16.73 & 79.15 & 41.43 & 92.19 & 43.1 & 18.49 & 78.42 & 59.0 & 51.09 & 79.9 & 58.88 & 31.54 \\

    Trans4PASS-T & SSL  & 51.86 & 78.24 & 41.16 & 85.82 & 27.86 & 36.01 & 30.92 & 21.26 & 17.70 & 79.11 & \textbf{46.44} & 93.47 & 44.72 & 17.66 & 79.44 & 63.69 & 48.14 & 81.56 & 59.09 & 32.96 \\
    Trans4PASS-T & MPA  & 51.93 & 77.27 & 45.61 & 85.66 & 23.57 & 37.10 & 31.22 & 20.13 & 15.35 & \textbf{79.91} & 43.81 & 93.95 & 46.37 & 21.63 & 79.34 & 62.09 & 56.05 & 78.43 & 56.31 & 32.89\\
    Trans4PASS-T & MPA + SSL  & 53.26 & 78.14 & 41.24 & 85.99 & 30.21 & 37.28 & 32.60 & 21.71 & 19.05 & 79.05 & 45.70 & 93.87 & 48.71 & 18.15 & 79.63 & 64.69 & 54.71 & 84.57 & 59.26 & 37.31\\
    \rowcolor{gray!15} Trans4PASS-T & MPA + SSL + MS  & \textbf{54.72} & 78.42 & 42.26 & 85.88 & 30.97 & 38.10 & 33.83 & 21.57 & \textbf{20.92} & 78.26 & 44.90 & 93.57 & 48.43 & 22.53 & 79.90 & 66.00 & 66.32 & 85.10 & 60.54 & 42.09\\
    \midrule 
    Trans4PASS-S & -  & 48.73 & 70.28 & 25.52 & 84.98 & 29.10 & 39.00 & 29.05 & 17.77 & 13.21 & 78.26 & 29.89 & 91.00 & 42.16 & 13.43 & 78.26 & 47.25 & 63.82 & 78.06 & 60.31 & 34.38 \\
    Trans4PASS-S & Warm-up  & 52.59 & 75.28 & 37.08 & 86.21 & 31.34 & 38.84 & 34.6 & 20.92 & 17.13 & 79.18 & 34.86 & 93.81 & 49.15 & 24.12 & 80.01 & 55.38 & 62.2 & 77.8 & 61.14 & 40.2 \\
    Trans4PASS-S & SSL  & 54.67 & 79.72 & 44.34 & 85.28 & 28.88 & 43.46 & 34.08 & 22.63 & 17.21 & 78.93 & 43.98 & 92.84 & 49.58 & \textbf{26.28} & 81.04 & 65.92 & 67.37 & 76.96 & 59.90 & 40.25 \\
    Trans4PASS-S & MPA  & 54.77 & \textbf{80.55} & \textbf{51.12} & \textbf{87.12} & 25.87 & \textbf{45.55} & 34.64 & 23.44 & 14.45 & 79.60 & 31.77 & \textbf{93.98} & 49.55 & 22.98 & 78.97 & 66.73 & 66.28 & 88.65 & 61.09 & 38.25\\
    Trans4PASS-S & MPA + SSL  & 55.25 & 78.39 & 41.62 & 86.47 & 31.56 & 45.47 & 34.02 & 22.98 & 18.33 & 79.63 & 41.35 & 93.80 & 49.02 & 22.99 & 81.05 & 67.43 & \textbf{69.64} & 86.04 & 60.85 & 39.20\\
    \rowcolor{gray!15} Trans4PASS-S & MPA + SSL + MS  & \textbf{56.38} & 79.91 & 42.68 & 86.26 & 30.68 & 42.32 & \textbf{36.61} & \textbf{24.81} & 19.64 & 78.80 & 44.73 & 93.84 & \textbf{50.71} & 24.39 & \textbf{81.72} & \textbf{68.86} & 66.18 & 88.62 & \textbf{63.87} & \textbf{46.62}\\
    \bottomrule[1pt]
    \end{tabular}}
  \end{center}
  \vskip-5ex
  \caption{\small \textbf{Per-class results on DensePASS dataset.} 
  `SSL' represents the self-supervised learning with pseudo-labels. `-' means no adaptation. `MS' denotes multi-scale evaluation.
  }
  \label{tab_sup:per_class_densepass}
  \vskip-3ex
\end{table*}

\begin{table*}[!t]
\renewcommand\arraystretch{0.95}
  \scriptsize
  \setlength{\tabcolsep}{7pt}
  \begin{center}
  \resizebox{0.99\textwidth}{!}{
    \begin{tabular}{l|c| c | c c c c c c c c c c c c c}
    \toprule
    \textbf{Network} & \textbf{Method} &  \textbf{\rotatebox{90}{mIoU}} &  \rotatebox{90}{beam} &  \rotatebox{90}{board} &  \rotatebox{90}{bookcase} & \rotatebox{90}{ceiling} &  \rotatebox{90}{chair} &  \rotatebox{90}{clutter} & \rotatebox{90}{column} &  \rotatebox{90}{door}&  \rotatebox{90}{floor} &  \rotatebox{90}{sofa} &  \rotatebox{90}{table} & \rotatebox{90}{wall} &  \rotatebox{90}{window} \\
    \toprule
    DANet & - & 40.28 & 0.00 & 56.07 & 52.09 & 72.05 & 35.72 & 20.54 & 5.81 & 19.43 & 72.84 & 31.76 & 41.80 & 68.43 & 47.13 \\
    DANet & P2PDA & 42.26 & 0.22 & 57.49 & 50.92 & 73.09 & 44.63 & 21.72 & 9.09 & 24.02 & 83.18 & 30.94 & 41.36 & 65.43 & 47.24 \\
    \midrule

    PVT-Tiny & - & 24.45 & 0.06 & 28.05 & 32.99 & 58.97 & 13.68 & 12.97 & 3.03 & 2.46 & 76.56 & 0.00 & 28.65 & 51.20 & 9.23 \\
    PVT-Tiny & P2PDA & 39.66 & 0.38 & 60.55 & 54.08 & 75.14 & 33.99 & 26.20 & 7.23 & 12.66 & 82.58 & 9.14 & 42.74 & 65.75 & 45.12 \\
    PVT-Small & - & 23.11 & 0.42 & 29.82 & 26.20 & 58.65 & 5.89 & 12.62 & 3.57 & 1.80 & 77.11 & 0.00 & 28.49 & 48.24 & 7.58 \\
    PVT-Small & P2PDA & 43.10 & 0.00 & 66.24 & 55.31 & 76.92 & 40.95 & 28.99 & 5.60 & 13.62 & 88.35 & 14.53 & 52.08 & 68.26 & 49.50 \\
    \midrule
    Trans4PASS-T &-& 46.08 & 0.28 & 65.21 & 60.07 & 76.36 & 50.30 & 33.09 & 11.89 & 20.72 & 86.87 & 26.14 & 50.84 & 68.64 & 48.56 \\
    \rowcolor{gray!15}Trans4PASS-T & MPA & 47.48 & 0.16 & 66.8 & 60.54 & 76.06 & 52.50 & 31.50 & 14.55 & 20.73 & 86.53 & 36.09 & 52.10 & 69.73 & 50.01\\
    Trans4PASS-S &-& 48.34 & \textbf{2.41} & \textbf{70.15} & 60.22 & 77.97 & \textbf{62.10} & \textbf{35.37} & 13.68 & 16.15 & 89.44 & 31.78 & \textbf{62.03} & 67.63 & \textbf{54.40} \\
    \rowcolor{gray!15}Trans4PASS-S & MPA  & \textbf{52.15} & 1.03 & 68.02 & \textbf{61.38} & \textbf{82.23} & 58.74 & 35.18 & \textbf{17.39} & \textbf{36.36} & \textbf{90.26} & \textbf{46.15} & 56.79 & \textbf{73.46} & 50.91 \\     \toprule
    DANet & supervised & 44.15 & 0.27 & 55.13 & 53.40 & 73.92 & 54.03 & \textbf{34.60} & 5.27 & 12.45 & 90.05 & 30.57 & 50.25 & 66.63 & 47.44 \\
    \rowcolor{gray!15}Trans4PASS-S & supervised & \textbf{53.31} & \textbf{0.43} & \textbf{69.45} & \textbf{62.24} & \textbf{82.77} & \textbf{58.52} & 34.26 & \textbf{21.86} & \textbf{44.87} & \textbf{91.19} & \textbf{40.78} & \textbf{57.69} & \textbf{74.80} & \textbf{54.20} \\
    \bottomrule
    \end{tabular}}
  \end{center}
  \vskip-5ex
  \caption{\small \textbf{Per-class results on Stanford2D3D-Panoramic dataset} according to the fold-1 data setting~\cite{stanford2d3d}.}
  \label{tab_sup:per_class_s2d3d_pan}
  \vskip-3ex
\end{table*}

\begin{table*}[!t]
\scriptsize
\renewcommand\arraystretch{0.75}
\setlength{\tabcolsep}{6pt}
\begin{center}
\resizebox{0.99\textwidth}{!}{
\begin{tabular}{c@{}|l|c| c | r r r r r r r r r r r r r}
    \toprule
    &\textbf{Method} & \textbf{Input} &  \textbf{\rotatebox{90}{mIoU}} &  \rotatebox{90}{beam} &  \rotatebox{90}{board} &  \rotatebox{90}{bookcase} & \rotatebox{90}{ceiling} &  \rotatebox{90}{chair} &  \rotatebox{90}{clutter} & \rotatebox{90}{column} &  \rotatebox{90}{door}&  \rotatebox{90}{floor} &  \rotatebox{90}{sofa} &  \rotatebox{90}{table} & \rotatebox{90}{wall} &  \rotatebox{90}{window} \\ \toprule
    {\multirow{17}{*}{\rotatebox[origin=c]{90}{\textit{Supervised}}}} 
    &StdConv~\cite{distortion_aware} &RGB& 32.6 &  0   & 46.6 & 44.9 & 60.8 & 32.4 & 18.8 & 0 & 13.0 & 78.0 & 0 & 32.6 & 54.8 & 40.1\\
    &CubeMap~\cite{distortion_aware} &RGB& 33.8 &  0.2 & 48.3 & 48.5 & 61.3 & 33.4 & 23.4 & 0 & 15.4 & 72.7 & 0 & 33.8 & 61.7 & 36.9\\
    &DistConv~\cite{distortion_aware} &RGB& 34.6 &  0.3 & 50.8 & 47.1 & 61.5 & 35.4 & 19.5 & 0 & 13.8 & 83.4 & 0 & 34.5 & 57.1 & 42.6\\
    &UNet~\cite{unet}&RGB-D& 35.9 & 8.5 & 27.2 & 30.7 & 78.6 & 35.3 & 28.8 & 4.9 & 33.8 & 89.1 & 8.2 & 38.5 & 58.8 & 23.9 \\
    &GaugeNet~\cite{gauge_equivariant} &RGB-D &39.4&--&--&--&--&--&--&--&--&--&--&--&--&-- \\
    &UGSCNN~\cite{spherical_unstructured_grids}& RGB-D  & 38.3 &  8.7 & 32.7 & 33.4 & 82.2 & 42.0 & 25.6 & 10.1 & 41.6 & 87.0 &  7.6 & 41.7 & 61.7 & 23.5\\
    &HexRUNet~\cite{orientation} &RGB-D & 43.3 & 10.9 & 39.7 & 37.2 & \textbf{84.8} & 50.5 & 29.2 & 11.5 & 45.3 & \textbf{92.9} & 19.1 & 49.1 & 63.8 & 29.4\\
    &Tangent (ResNet-101)~\cite{tangent} & RGB  & 45.6 &--&--&--&--&--&--&--&--&--&--&--&--&--\\
    &HoHoNet (ResNet-101)~\cite{hohonet} & RGB  & 52.0 &--&--&--&--&--&--&--&--&--&--&--&--&--\\
    & \grow{Trans4PASS (F-1)} & \grow{RGB} & \grow{53.3} & \grow{0.4} & \grow{69.5} & \grow{62.2} & \grow{82.8} & \grow{58.5} & \grow{34.3} & \grow{21.9} & \grow{44.9} & \grow{91.2} & \grow{40.8} & \grow{57.7} & \grow{74.8} & \grow{54.2}\\
    & \grow{Trans4PASS (F-2)} & \grow{RGB} & \grow{45.7} & \grow{12.5} & \grow{46.9} & \grow{32.6} & \grow{82.3} & \grow{64.7} & \grow{37.5} & \grow{20.1} & \grow{42.7} & \grow{86.6} & \grow{17.7} & \grow{45.2} & \grow{70.3} & \grow{35.1}\\
    & \grow{Trans4PASS (F-3)} & \grow{RGB} & \grow{57.2} & \grow{21.4} & \grow{65.4} & \grow{58.3} & \grow{80.2} & \grow{55.8} & \grow{41.9} & \grow{28.6} & \grow{76.3} & \grow{88.6} & \grow{45.4} & \grow{58.8} & \grow{59.3} & \grow{63.6}\\
    &Trans4PASS (Avg) & RGB& 52.1 & \textbf{11.4} & 60.6 & 51.1 & 81.8 & 59.7 & 37.9 & 23.5 & \textbf{54.6} & 88.8 & \textbf{34.6} & 53.9 & 68.1 & 51.0\\
    & \grow{Trans4PASS (F-1, MS)} & \grow{RGB} & \grow{54.2} & \grow{0.7} & \grow{72.1} & \grow{64.1} & \grow{83.4} & \grow{61.3} & \grow{35.5} & \grow{22.4} & \grow{42.2} & \grow{92.0} & \grow{41.6} & \grow{59.4} & \grow{75.3} & \grow{54.4}\\
    & \grow{Trans4PASS (F-2, MS)} & \grow{RGB} & \grow{46.4} & \grow{3.1} & \grow{48.2} & \grow{32.1} & \grow{82.9} & \grow{66.4} & \grow{37.8} & \grow{20.3} & \grow{42.7} & \grow{87.2} & \grow{16.8} & \grow{45.9} & \grow{71.3} & \grow{38.0}\\
    & \grow{Trans4PASS (F-3, MS)} & \grow{RGB} & \grow{58.4} & \grow{1.7} & \grow{67.1} & \grow{60.1} & \grow{81.3} & \grow{56.8} & \grow{42.6} & \grow{29.8} & \grow{77.6} & \grow{89.5} & \grow{45.3} & \grow{59.9} & \grow{60.1} & \grow{67.3}\\
    \rowcolor{gray!15}\cellcolor{white} &Trans4PASS (Avg, MS) & RGB& \textbf{53.0} & 1.8 & \textbf{62.5} & \textbf{52.1} & 82.6 & \textbf{61.5} & \textbf{38.6} & \textbf{24.2} & 54.2 & 89.5 & 34.5 & \textbf{55.1} & \textbf{68.9} & \textbf{53.2}\\
    \midrule
    {\multirow{12}{*}{\rotatebox[origin=c]{90}{\textit{UDA}}}}
    & \grow{Trans4PASS (F-1)} & \grow{RGB} & \grow{48.6} & \grow{0.1} & \grow{65.8} & \grow{58.3} & \grow{80.5} & \grow{54.2} & \grow{29.1} & \grow{17.4} & \grow{23.7} & \grow{89.0} & \grow{34.3} & \grow{54.9} & \grow{73.2} & \grow{51.6}\\
    & \grow{Trans4PASS (F-2)} & \grow{RGB} & \grow{40.6} & \grow{10.2} & \grow{38.3} & \grow{28.9} & \grow{77.8} & \grow{54.6} & \grow{32.5} & \grow{15.7} & \grow{32.9} & \grow{83.2} & \grow{13.7} & \grow{38.0} & \grow{67.9} & \grow{33.6}\\
    & \grow{Trans4PASS (F-3)} & \grow{RGB} & \grow{55.2} & \grow{17.4} & \grow{64.7} & \grow{60.2} & \grow{76.4} & \grow{58.3} & \grow{41.4} & \grow{5.0} & \grow{76.6} & \grow{84.5} & \grow{47.2} & \grow{57.3} & \grow{63.8} & \grow{64.5}\\
    &Trans4PASS (Avg) & RGB & 48.1 & 9.2 & 56.3 & 49.1 & 78.2 & 55.7 & 34.3 & \textbf{12.7} & 44.4 & 85.6 & 31.8 & 50.1 & 68.3 & 49.9\\ 
    & \grow{Trans4PASS (F-1, MPA)} & \grow{RGB} & \grow{52.2} & \grow{1.0} & \grow{68.0} & \grow{61.4} & \grow{82.2} & \grow{58.7} & \grow{35.2} & \grow{17.4} & \grow{36.4} & \grow{90.3} & \grow{46.2} & \grow{56.8} & \grow{73.5} & \grow{50.9}\\
    & \grow{Trans4PASS (F-2, MPA)} & \grow{RGB} & \grow{41.8} & \grow{11.0} & \grow{35.1} & \grow{30.9} & \grow{78.6} & \grow{59.3} & \grow{32.7} & \grow{14.3} & \grow{45.6} & \grow{80.1} & \grow{22.9} & \grow{37.0} & \grow{66.2} & \grow{29.6}\\
    & \grow{Trans4PASS (F-3, MPA)} & \grow{RGB} & \grow{58.5} & \grow{24.5} & \grow{70.4} & \grow{59.0} & \grow{81.3} & \grow{58.5} & \grow{43.3} & \grow{4.6} & \grow{76.1} & \grow{89.6} & \grow{53.3} & \grow{62.0} & \grow{65.7} & \grow{72.0}\\
    &Trans4PASS (Avg, MPA) & RGB & 50.8 & \textbf{12.2} & 57.8 & 50.4 & 80.7 & 58.8 & 37.1 & 12.1 & \textbf{52.7} & \textbf{86.7} & 40.8 & 51.9 & \textbf{68.4} & 50.8\\
    & \grow{Trans4PASS (F-1, MPA, MS)} & \grow{RGB} & \grow{52.6} & \grow{0.8} & \grow{70.7} & \grow{63.3} & \grow{82.2} & \grow{60.8} & \grow{36.2} & \grow{16.4} & \grow{33.4} & \grow{90.5} & \grow{45.9} & \grow{58.4} & \grow{73.1} & \grow{51.5}\\
    & \grow{Trans4PASS (F-2, MPA, MS)} & \grow{RGB} & \grow{42.6} & \grow{11.7} & \grow{35.5} & \grow{31.6} & \grow{79.2} & \grow{60.8} & \grow{33.2} & \grow{15.6} & \grow{46.5} & \grow{78.8} & \grow{24.1} & \grow{38.0} & \grow{66.2} & \grow{32.5}\\
    & \grow{Trans4PASS (F-3, MPA, MS)} & \grow{RGB} & \grow{58.3} & \grow{22.6} & \grow{70.6} & \grow{59.4} & \grow{81.5} & \grow{58.8} & \grow{43.9} & \grow{4.2} & \grow{76.7} & \grow{89.5} & \grow{52.8} & \grow{62.0} & \grow{66.0} & \grow{70.7}\\
    \rowcolor{gray!15}\cellcolor{white} &Trans4PASS (Avg, MPA, MS) & RGB & \textbf{51.2} & {11.7} & \textbf{58.9} & \textbf{51.4} & \textbf{81.0} & \textbf{60.1} & \textbf{37.7} & 12.0 & {52.2} & {86.2} & \textbf{40.9} & \textbf{52.8} & \textbf{68.4} & \textbf{51.6}\\
    \bottomrule
\end{tabular}}
\end{center}
\vskip-5ex
\caption{\small \textbf{Comparison on Stanford2D3D-Panoramic dataset.} `F-$i$' is the result of the fold-$i$ (in \grow{gray}) setting of Stanford2D3D~\cite{stanford2d3d}. `Avg' is the averaged result of all $3$ folds. 
`MS' is multi-scale evaluation. `UDA' is short for unsupervised domain adaptation.}
\label{tab_sup:supervised_s2d3d_pan}
\vskip-4ex
\end{table*}

\begin{figure*}[t]
	\begin{subfigure}{\textwidth}
    \includegraphics[width=0.95\textwidth]{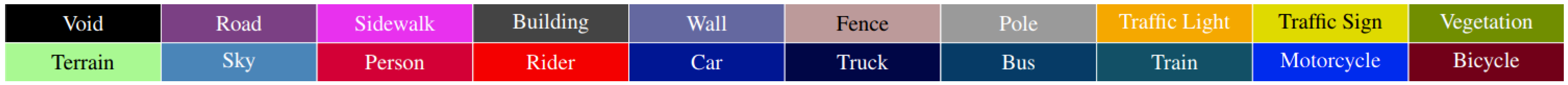}
    \centering \vskip -3ex
    \end{subfigure}
	\begin{subfigure}{\textwidth}
	\centering
    \includegraphics[width=0.95\textwidth]{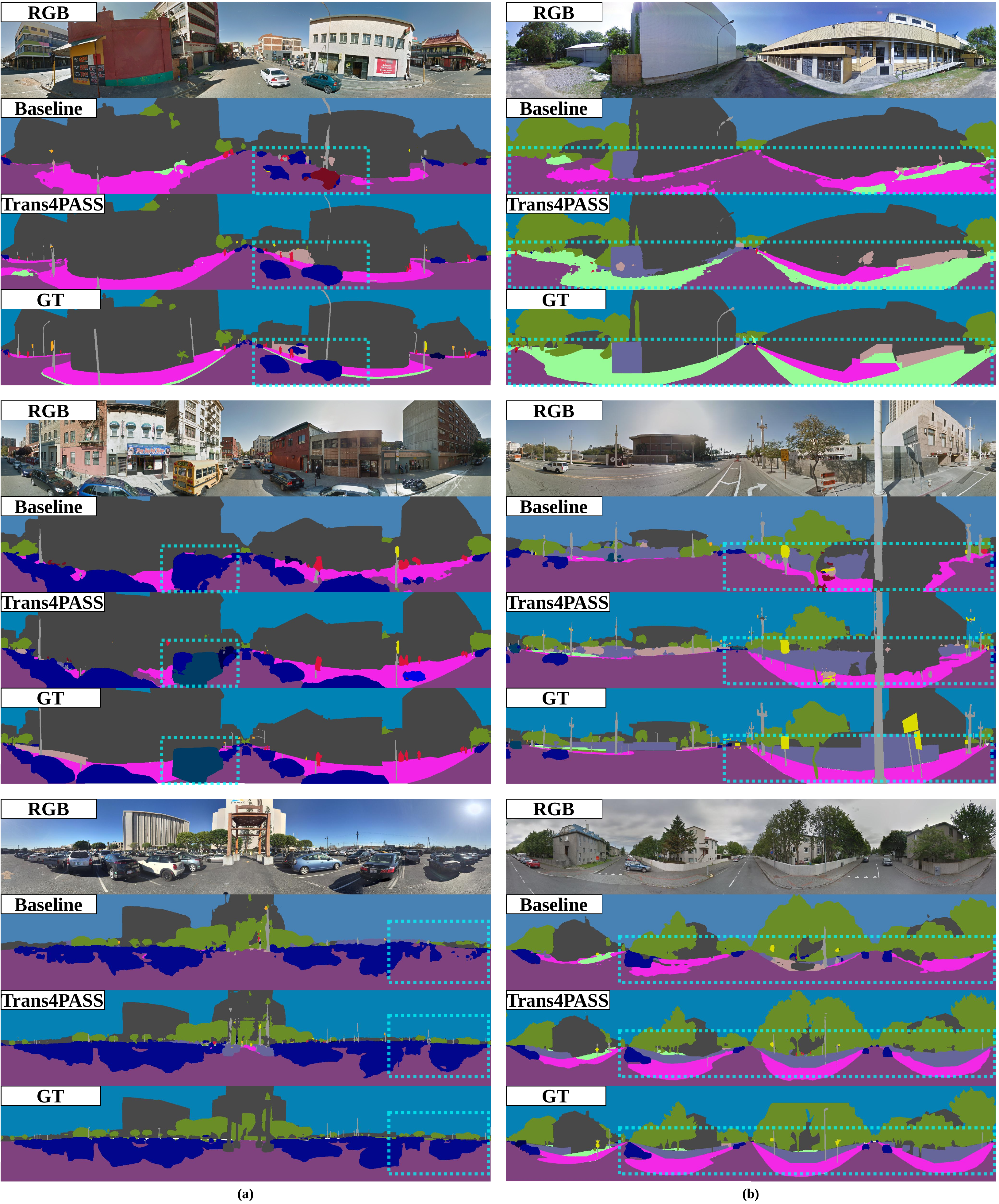}
    \end{subfigure}
    \vskip -1ex
	\caption{\textbf{Qualitative comparisons} in outdoor scenarios.} 
	\label{fig:supple_vis_outdoor}
\end{figure*}

\subsection{More visualizations in outdoor scenarios}
To fully demonstrate the effect of Trans4PASS in dealing with image distortions and object deformations, more qualitative comparisons between the baseline and the proposed Trans4PASS are displayed in Fig.~\ref{fig:supple_vis_outdoor}, which are generated from the evaluation set of DensePASS dataset~\cite{densepass}. Specifically, Trans4PASS can better classify and segment deformed foreground objects with accurate boundaries, such as the segmentation results of \emph{cars} and \emph{trucks} highlighted by the blue dashed rectangles in Fig.~\ref{fig:supple_vis_outdoor}(a), while the baseline model without deformable PE and deformable MLP modules is likely to be confused or fail in these categories. Apart from the foreground object, the ultra-wide arranged background is particularly distorted and challenging. Thanks to the two distortion-aware modules, our Trans4PASS yields high-quality segmentation results in these categories, \eg, \emph{terrain}, \emph{sidewalk}, and \emph{wall} in Fig.~\ref{fig:supple_vis_outdoor}(b).

\section{Broader Impact.} 
This work promotes panoramic semantic segmentation of indoor and outdoor scenes, which benefits ultra-wide scene understanding. However, the proposed method has not been verified in practical applications such as those in intelligent vehicles and mobility assistive systems. As the experiments are conducted based on the referred datasets, there are still data biases in different test fields. If the learned model is directly applied to real scenarios, it may cause negative social impacts such as less reliable decision with less accurate segmentation, which should be considered in the downstream applications. 
\end{document}